%% file: main.tex
\pdfoutput=1

\documentclass[11pt]{article}

\input{header/header}

\title{Confidence Calibration in Large Language Model-Based Entity Matching}

\author{
\textbf{Iris Kamsteeg \textsuperscript{1}},
\textbf{Juan Cardenas-Cartagena\textsuperscript{1}},
\textbf{Floris van Beers\textsuperscript{2}},
\textbf{Gineke ten Holt\textsuperscript{2}},\\
\textbf{Tsegaye Misikir Tashu\textsuperscript{1}},
\textbf{Matias Valdenegro-Toro \textsuperscript{1}}
\vspace{0.5em}
\\
 \textsuperscript{1}Bernoulli Institute, University of Groningen, The Netherlands,
 \textsuperscript{2}Independent Researcher
\\
 \texttt{
   \href{mailto:ikamsteeg@ziggo.nl}{ikamsteeg@ziggo.nl}, \href{mailto:t.m.tashu@rug.nl}{t.m.tashu@rug.nl}, \href{mailto:m.a.valdenegro.toro@rug.nl}{m.a.valdenegro.toro@rug.nl}
 }
}

\begin{document}
\maketitle

\begin{abstract}

\input{sections/abstract}
\end{abstract}

\section{Introduction}
\label{sec:introduction}

\input{sections/introduction}

\section{Confidence Calibration}
\label{sec:confidence_calibration}
\input{sections/confidence_calibration}

\section{Related Work}
\label{sec:related_work}

\input{sections/related_work}

\section{Methods}
\label{sec:methods}
\input{sections/methods}

\section{Results}
\label{sec:results}

\input{sections/results}

\section{Conclusions}
\label{sec:conclusion}
\input{sections/conclusion}

\label{sec:future_work}

\input{sections/future_work}

\section*{Limitations}
\label{sec:limitations}
\input{sections/limitations}

\section*{Acknowledgements}
We thank the Center for Information Technology of the University of Groningen for their support and for providing access to the Hábrók high performance computing cluster. 

\section*{Broader Impact Statement}

\input{sections/ethics_statement}
\bibliography{anthology,custom}

\appendix

\input{sections/appendix}

\end{document}

%% file: header/header.tex
\usepackage[final]{acl}

\usepackage{times}
\usepackage{latexsym}

\usepackage[T1]{fontenc}
\usepackage[utf8]{inputenc}

\usepackage{microtype}

\usepackage{amsmath}
\newcommand*{\fullref}[1]{\hyperref[{#1}]{\autoref*{#1} \nameref*{#1}}}
\usepackage{graphicx}
\usepackage{natbib}
\usepackage[table]{xcolor} %
\usepackage{xcolor}
\usepackage{caption}
\usepackage{subcaption}

%% file: sections/abstract.tex
This research aims to explore the intersection of Large Language Models and confidence calibration in Entity Matching. To this end, we perform an empirical study to compare baseline RoBERTa confidences for an Entity Matching task against confidences that are calibrated using Temperature Scaling, Monte Carlo Dropout and Ensembles. We use the Abt-Buy, DBLP-ACM, iTunes-Amazon and Company datasets. The findings indicate that the proposed modified RoBERTa model exhibits a slight overconfidence, with Expected Calibration Error scores ranging from 0.0043 to 0.0552 across datasets. We find that this overconfidence can be mitigated using Temperature Scaling, reducing Expected Calibration Error scores by up to 23.83\%. 

%% file: sections/introduction.tex
Entity Resolution (ER) can be defined as the task of determining which data entries across different data sources refer to the same real-world entity. A key sub-task of ER is Entity Matching (EM), which specifically addresses the binary classification problem of determining whether pairs of data entries from different sources refer to the same entity \citep{christophides_overview_2020}. In today’s data-driven era, EM plays a critical role in various domains, including the medical field \citep{jaro_probabilistic_1995, meray_probabilistic_2007}, where accurate matching can improve patient care; the reconstruction of historical populations by linking birth, marriage, and death records \citep{bloothooft_population_2015}; and law enforcement, where matching data entries is vital for investigations and crime prevention \citep{dahlin2012combining}.

The state-of-the-art methods for performing EM utilize Transformer-based architectures \citep{vaswani_attention_2017}, pre-trained Large Language Models (LLMs) \citep{brunner_entity_2020, li_deep_2020, peeters_intermediate_2020, peeters_dual-objective_2021, peeters_supervised_2022, peeters_using_2023, peeters_entity_2024}, such as RoBERTa \citep{liu_roberta_2019} and GPT-4 \citep{openai_gpt-4_2024}.

\begin{figure}[]
    \centering
    \includegraphics[width=0.18\textwidth]{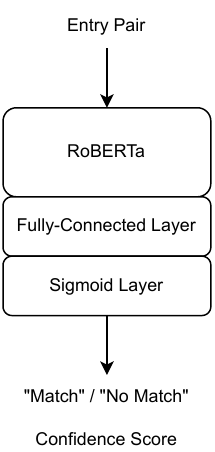}
    \caption{Overview of this research's model (without any confidence calibration methods visualised), model input and model output. In addition to classifying each entry pair as a `match’ or `no match,’ the model also generates a score that should reflect the model's confidence in its prediction.}
    \label{fig:overview}
\end{figure}
 
However, while these models are successful, they have, in other Natural Language Processing tasks, shown to struggle to accurately express their confidence in predictions and can exhibit overconfidence (\citealp{desai_calibration_2020, jiang_how_2021}). Ideally, a model provides information about its certainty alongside its predictions. For example, in a binary EM task, a model would output a `match' or `no match' prediction label alongside a probability, or confidence score, that is reliable. Refining models' predicted confidence scores to ensure that they accurately reflect the true likelihood of the predictions is called confidence calibration. While the topic of confidence calibration on LLMs has been explored (\citealp{desai_calibration_2020, sankararaman_bayesformer_2022, chen_calibrating_2024}), the intersection of confidence calibration, LLMs, and the application of EM has not yet been researched. Yet, confidence calibration is important as it provides transparency over models' results \citep{ghahramani_probabilistic_2015}. For example, the distribution of confidence in a model's EM predictions can give the user insights into the model's overall reliability in the task. Precise confidence scores can also play a crucial role in guiding subsequent tasks. Moreover, confidence scores can be used to help researchers better understand a model's inner workings. Finally, they can help in improving a model: when it is clear in what specific cases a model is uncertain, it is easier to see a model's weak points and with that, possible areas for improvement.

\textbf{Contributions}. This paper aims to explore the confidence calibration performance of LLMs in EM and benchmark confidence calibration methods to enhance their performance. We focus on pretrained RoBERTa \citep{liu_roberta_2019} as the LLM of interest as it has a competitive performance among LLM models for EM \citep{li_deep_2020, peeters_dual-objective_2021, peeters_entity_2024}. In contrast to other state-of-the-earth models for EM, RoBERTa is open-source and lightweight. Our study assesses the confidence calibration performance in EM using the Expected Calibration Error (ECE) as the primary metric. We evaluate fine-tuned RoBERTa model's ECE scores both with and without the use of confidence calibration methods and investigate which methods yield the greatest improvement. Since confidence calibration methods may influence the model’s predictions, we additionally examine their effect on the $F_1$ score to ensure that improved confidence calibration does not come at the cost of classification performance. Furthermore, we analyze confidence histograms, reliability diagrams, the Maximum Calibration Error (MCE) and the Root Mean Square Calibration Error (RMSCE). The confidence calibration methods tested are Temperature Scaling \citep{guo_calibration_2017}, Monte Carlo Dropout \citep{gal_dropout_2016}, and Ensembles \citep{lakshminarayanan_simple_2017}. We use the Abt-Buy, DBLP-ACM (dirty and structured) \citep{kopcke_evaluation_2010}, iTunes-Amazon (dirty and structured), and Company  \citep{konda_magellan_2016} datasets, ensuring diversity in terms of data content, size and structure.

Figure \ref{fig:overview} presents an overview of the proposed modified RoBERTa model used in this research. As shown, the goal is to obtain confidence scores that accurately reflect the model's confidence in its EM predictions. Confidence calibration methods can help improve these scores. 

%% file: sections/confidence_calibration.tex
We say that a model is well-calibrated if its prediction's confidence scores accurately reflect the probability of those predictions being correct. For EM, for example, all pairs that are predicted to match with around 0.5 to 0.6 confidence  should be actual matching pairs 50 to 60\% of the time. This is also referred to as the alignment between the `predicted probability' (the confidence) and the `empirical probability' \citep{naeini_obtaining_2015, guo_calibration_2017, kuppers_confidence_2022}. Generally, for a binary classification task such as EM, the `confidence' signifies the confidence of a prediction belonging to the positive class (in the case of EM: a `match'). The predicted probability of the positive class then needs to align with the empirical probability of the positive class. `High confidences', in this context, generally denote predicted probabilities close to either 0 or 1, while `low confidences' denote predicted probabilities close to 0.5. 

The confidence calibration of models has been evaluated by plotting confidence histograms and reliability diagrams, and by measuring the Expected Calibration Error (ECE) \citep{naeini_obtaining_2015} or similar metrics such as the Maximum Calibration Error (MCE) \citep{naeini_obtaining_2015} and Root Mean Square Calibration Error (RMSCE) \citep{kumar_verified_2019}. Intuitively, these scores measure the difference between the predicted probability and the empirical probability, and should therefore be minimized to optimize the confidence calibration. Compared to the ECE, the MCE is useful in production settings where reliable confidence measures are absolutely necessary due to high risks. This is due to its measure of the worst-case deviation between the predicted probabilities and the empirical probabilities. When comparing the ECE to the RMSCE, the latter places a greater emphasis on larger errors.

%% file: sections/related_work.tex
\subsection{Large Language Models for Entity Matching}

Various pre-trained LLMs have shown state-of-the-art results for EM tasks. \citet{brunner_entity_2020}, for example, analysed the performance of four LLMs: BERT \citep{devlin_bert_2019}, XLNet \citep{yang_xlnet_2019}, RoBERTa \citep{liu_roberta_2019} and DistilBERT \citep{sanh_distilbert_2020}, and found an increase in $F_1$ scores of up to 35.9\% compared to state-of-the-art non-LLM methods. Other state-of-the-art results were presented by \citet{li_deep_2020}, who introduced DITTO: an EM system that combines the use of LLMs such as BERT, DistilBERT and RoBERTa with various optimisation techniques; and \citet{peeters_intermediate_2020, peeters_dual-objective_2021, peeters_supervised_2022}, who experimented with BERT and RoBERTa-SupCon for EM in the product domain.

Decoder-only models have more recently caught the attention in the field. \citet{narayan_can_2022} compared GPT-3 against the DITTO system. The performance of GPT-3 \citep{brown_language_2020} using few-shot learning was better than DITTO's performance for four out of seven datasets. In their paper ``Using ChatGPT for Entity Matching'', \citet{peeters_using_2023} test the performance of ChatGPT (GPT3.5) on an EM task using product data. They find that though the results of ChatGPT on this data is generally worse compared to the results of a finetuned RoBERTa, it is beneficial that ChatGPT does not necessarily require any finetuning, and, thus, performs well on unseen data. Peeter and Bizers' study ``Entity Matching using Large Language Models'' \citep{peeters_entity_2024} shows that GPT-4 \citep{openai_gpt-4_2024} especially performs well in EM tasks.

\subsubsection{Confidence Calibration of Large Language Models}

While in the early 2000s, simple neural networks typically produced well-calibrated probabilities in binary classification tasks \citep{niculescu-mizil_predicting_2005}, recent studies have shown that this is generally not the case for more modern neural networks. In their 2017 paper ``On the Calibration of Modern Neural Networks'' \citep{guo_calibration_2017}, Guo et al. showed that state-of-the-art neural networks of that time (including ResNet \cite{he_deep_2016}), do not show a good confidence calibration at all. The researchers also indicate that miscalibration worsens as the classification error is reduced. \citet{desai_calibration_2020}, as well as \citet{xiao_uncertainty_2022} explored the confidence calibration of BERT \citep{devlin_bert_2019} and RoBERTa \citep{liu_roberta_2019} in natural language inference, paraphrase detection, sentiment analysis and commonsense reasoning tasks. While BERT and RoBERTa show less miscalibration than the models that were evaluated by Guo and colleagues, the confidence calibration of the LLMs does show room for improvement. In a study by \citet{jiang_how_2021}, decoder-only LLMs were also shown to be generally miscalibrated and often overconfident \citep{jiang_how_2021}. 

One of the reasons that LLMs do not seem to produce well-calibrated predictions is that they are not trained to do this as an explicit learning goal. Instead, during training, these networks are encouraged to assign high confidences, in the form of sigmoid scores, to the correct class, without regard to nuances that prediction probabilities should ideally have \citep{hendrycks2017a}. 

However, various methods have been introduced to improve the confidence calibration of LLMs. These include Temperature Scaling \citep{guo_calibration_2017}, Monte Carlo Dropout \citep{gal_dropout_2016} and Ensembles \cite{lakshminarayanan_simple_2017}.

%% file: sections/methods.tex
\subsection{Data}

Six datasets are used in this study: Abt-Buy, DBLP-ACM-Structured, DBLP-ACM-Dirty \citep{kopcke_evaluation_2010}, iTunes-Amazon-Structured, iTunes-Amazon-Dirty, and Company \citep{konda_magellan_2016}. For DBLP-ACM and iTunes-Amazon, the structured and dirty versions of the datasets contain the same entries, but for the dirty version, there is a 50\% chance that an attribute value is moved to a different attribute. Table \ref{tab:experiment_1_data_splits} presents the domains of the datasets, as well as the number of pairs for each dataset, for each split. In brackets is the percentage of positive pairs.

\begin{table*}[h]
\centering
\begin{tabular}{lllll}
\textbf{Dataset name} & \textbf{Domain} & \textbf{Training pairs} & \textbf{Validation pairs}                                          & \textbf{Testing pairs} \\ \hline
\textbf{Abt-Buy} & Products & 5743 (10.72\%) & 1916 (10.75\%) & 1916 (10.75\%) \\
\textbf{DBLP-ACM}* & Citations & 7417 (17.96\%) & 2473 (17.96\%) & 2473 (17.96\%)\\
\textbf{iTunes-Amazon}* & Songs & 321 (24.30\%) & 109 (27.78\%) & 109 (27.78\%)\\
\textbf{Company} & Companies & 67 596 (24.94\%) & 22 533 (25.30\%) & 22 503 (25.06\%) \\
\hline
\end{tabular}
\caption{Overview of the dataset's domains and data splits, along with the percentage of positive pairs per split between brackets. *The splits and percentages are the same for both the structured and dirty versions.}
\label{tab:experiment_1_data_splits}
\end{table*}

\subsection{Model}

We use RoBERTa \citep{liu_roberta_2019}, pretrained on English language, as target LLM for EM. RoBERTa was one of the first LLMs to be used for EM and performs among the best of all tested non-decoder LLMs for EM, while not using any additional optimisation techniques \citep{brunner_entity_2020, li_deep_2020}. We utilise Huggingface's pre-trained RoBERTa base model\footnote{https://huggingface.co/FacebookAI/roberta-base}.

We adopt the setup by \citet{li_deep_2020} to make RoBERTa suitable for EM in the proposed datasets. That is, a single fully connected layer and sigmoid output layer are added after the final layer of the pre-trained RoBERTa base model. These two added layers, together with the RoBERTa base model, constitute the EM model. The fully connected layer's parameters are randomly initialized. The RoBERTa EM model is fed pairs of entries and outputs whether or not the pairs of entries are predicted as a `match' (label 1) or `no match' (label 0). We adopt \citet{li_deep_2020} method of data serialization to convert structured EM data into sequences of text that can be fed to the RoBERTa model. Hyper-parameters are also taken from the paper of \citet{li_deep_2020}.

In order for the model to understand the task and the data that it is given, fine-tuning is performed on the RoBERTa base model along with the fully connected and sigmoid layers using supervised training with a binary cross-entropy loss

\subsection{Confidence Calibration Methods}

\subsubsection{Temperature Scaling}

Temperature Scaling was introduced by \citet{guo_calibration_2017} as a single-parameter version of Platt Scaling \citep{platt_probabilistic_1999}. The method is easy to realise and understand, and is time-efficient and lightweight. It has led to improvements in confidence calibration for both encoder-only and decoder-only LLMs for sentiment analysis, natural language inference, common sense reasoning, paraphrase detection, and question-answering tasks. For some datasets and tasks, the technique has resulted in ECEs that are up to ten times smaller compared to those of uncalibrated models \citep{guo_calibration_2017, desai_calibration_2020, jiang_how_2021, xiao_uncertainty_2022}.

\subsubsection{Monte Carlo Dropout}

Monte Carlo Dropout was introduced by \citet{gal_dropout_2016} and applies dropout with probability $p$ \citep{hinton_improving_2012} at inference time. It has shown to, with its regularizing effect, improve the confidence calibration of models in tasks such as sentiment analysis, natural language inference, commonsense reasoning, named entity recognition and language modeling \citep{xiao_quantifying_2019, xiao_uncertainty_2022}. 

In our implementation, dropout is applied to the fully connected layer of the EM model. We perform dropout for just this layer to make the confidence calibration method implementation as lightweight as possible. 

\subsubsection{Ensembles}

Ensembles can be used for confidence calibration by separately training multiple instances of a model and using the mean probability outputs at inference time \citep{lakshminarayanan_simple_2017}. Through their regularizing effect, Ensembles have shown to improve the confidence calibration across various tasks including sentiment analysis, natural language inference and commonsense reasoning \citep{xiao_uncertainty_2022}.

We apply Ensembles on the fully connected layer and the sigmoid activation layer. In this way, we minimize the number of times that entry pairs need to pass through the RoBERTa base model.

\subsubsection{Experimental Setup}

First, the performance of the baseline RoBERTa EM model is evaluated in terms of $F_1$ score and confidence calibration for all datasets. To this end, we train and test on five independently randomly initialized RoBERTa EM models. For each run, the training data are shuffled. We adopt the number of epochs specified in the code by \citet{li_deep_2020} for all datasets. This corresponds to 40 epochs. The model checkpoint that generates the highest $F_1$ score on the validation set is used for testing. The sigmoid scores that the model produces for the testing set are used as baseline predicted probabilities. 

Secondly, Temperature Scaling, Monte Carlo Dropout, and Ensembles are individually applied and evaluated. They are compared against each other and against the baseline.

In applying Temperature Scaling, we adopt the approach by \citet{mukhoti_calibrating_2020} to find the best values for the temperature $T$. We use a similar approach to find the dropout value $p$ for the Monte Carlo Dropout method. For each dataset and experiment run, $T$ and $p$ are determined by minimizing the ECE on the validation set through a single parameter grid-search, while avoiding any decrease in the $F_1$ score. 

For Temperature Scaling, we take, for each trained RoBERTa model (i.e. one model per run per dataset), the sigmoid scores on the validation set. These are scaled with temperatures $T \in \{0.1, 0.2, 0.3, 0.4, ..., 9.9, 10.0\}$. Next, the ECE is calculated over all of the scaled sigmoid scores. For each dataset and run, the $T$ is recorded that results in the smallest ECE on the validation set. Next, these values for $T$ are used on the corresponding testing set sigmoid scores. The final results consist of the temperatures and, most importantly, the ECEs of the test sets. Note that Temperature Scaling does not change the $F_1$ scores.

For Monte Carlo Dropout, we take the best RoBERTa EM models from previous experiments for each dataset and run, and apply Monte-Carlo Dropout with $p \in \{0.05, 0.10, 0.15, 0.20, ..., 0.90, 0.95\}$. For each dataset, experiment run and dropout value, the model predicts over the validation set ten times. The resulting sigmoid scores from these ten sub-runs are averaged using the mean. 

For all averaged sigmoid scores, the $F_1$ score and ECE are calculated. For each dataset and run, the $p$ is recorded that results in the smallest ECE on the validation set, while maintaining an $F_1$ score not lower than the original score without dropout. If all values of $p$ decrease the $F_1$ score, a dropout value of 0.00 is recorded. 

Next, for each dataset and run, these recorded values for $p$ are used while performing inference on the corresponding test sets. Inference is performed ten times for each dataset and run using the recorded dropout probabilities. Afterwards, the means of the resulting sigmoid scores are calculated, and the $F_1$ scores and ECEs are computed over these means.

For Ensembles, for each dataset and experiment run, we randomly initialise the fully connected layer weights five times. For each dataset and experiment run, we then train, validate and test, using these five differently initialised models. After doing this, we compute the means over the five ensemble runs' test sets sigmoid scores. These average sigmoid scores are then used to derive the final $F_1$ scores and ECEs.

Evaluation for the baseline RoBERTa EM model and the confidence calibration methods occurs in terms of confidence histograms, reliability diagrams, $F_1$ score, ECE, MCE, and RMSCE metrics, using a number of bins = $\sqrt{|\mathcal{D}|}$, with $\mathcal{D}$ being the dataset. A paired t-test is used to assess the significance of differences between the baseline results for the Temperature Scaling and Monte Carlo Dropout methods. An unpaired t-test is used to do the same for the Ensembles method.

%% file: sections/results.tex
Table \ref{tab:calibration_results} presents the mean $F_1$ scores, ECEs, MCEs, and RMSCEs of various confidence calibration methods, over five runs, for all  datasets. It also presents the baseline confidence calibration using the RoBERTa sigmoid scores without any confidence calibration method applied. 

Appendix \ref{sec:roberta_em_performance} presents a more detailed overview of the performance of the baseline RoBERTa model in terms of $F_1$ score, precision, recall and inference time.

\begin{table*}[t]
\centering
\vspace{-16pt}
\begin{tabular}{l|llll}
\textbf{Dataset}                &  \textbf{ECE} & $\mathbf{F_1}$ & \textbf{MCE} & \textbf{RMSCE} \\ \hline
\multicolumn{5}{c}{\textbf{Baseline}}        \\ \hline
Abt-Buy & 0.0193
 ± 0.0018
 & 90.81
 ± 0.85
 & 0.9305
 ± 0.0469
 & 0.0558
 ± 0.0032
 \\ 
        DBLP-ACM-S & 0.0041
 ± 0.0010
 & 98.78

 ± 0.40

 & 0.7800
 ± 0.2900
 & 0.0303
 ± 0.0131
 \\ 
        DBLP-ACM-D & 0.0043
 ± 0.0011
 & 98.85 ± 0.18
 & 0.6949
 ± 0.1204
 & 0.0287
 ± 0.0104
 \\ 
        iTunes-Amazon-S & 0.0391
 ± 0.0064
 & 90.53
 ± 1.64
 & 0.3085
 ± 0.2024
 & 0.0506
 ± 0.0113
 \\ 
        iTunes-Amazon-D & 0.0410
 ± 0.0121
 & 91.50
 ± 1.90
 & 0.3460
 ± 0.2285
 & 0.0683
 ± 0.0181
\\ 
        Company & 0.0552
 ± 0.0099
 & 82.75
 ± 0.92
 & 0.5449
 ± 0.0855
 & 0.0967
 ± 0.0177
 \\ 
        \hline
\multicolumn{5}{c}{\textbf{Temperature Scaling}}
\\ \hline Abt-Buy & \cellcolor{green!45}$^\downarrow$0.0147
 ± 0.0017
 & 90.81 ± 0.85 & \cellcolor{green!45}$^\downarrow$0.8539
 ± 0.0882
 & \cellcolor{red!45}$^\uparrow$0.0632
 ± 0.0046
 \\ 
         DBLP-ACM-S & \cellcolor{green!45}$^\downarrow$0.0036
 ± 0.0011
 & 98.78 ± 0.40 & \cellcolor{red!10}0.7580
 ± 0.2031
 & \cellcolor{red!10}0.0306
 ± 0.0087
 \\ 
        DBLP-ACM-D & \cellcolor{green!10}0.0038

 ± 0.0011

 & 98.85 ± 0.18 & \cellcolor{red!10}0.7983
 ± 0.2174
 & \cellcolor{red!45}$^\uparrow$0.0312
 ± 0.0085
 \\ 
        iTunes-Amazon-S & \cellcolor{green!45}$^\downarrow$0.0352
 ± 0.0118
 & 90.53 ± 1.63 & \cellcolor{red!10}0.3394
 ± 0.2089
 & \cellcolor{green!10}0.0415
 ± 0.0226
 \\ 
        iTunes-Amazon-D & \cellcolor{green!10}0.0377
 ± 0.0102
 & 91.50 ± 1.90 & \cellcolor{red!10}0.4036
 ± 0.3247
 & \cellcolor{green!10}0.0649
± 0.0288
\\ 
        Company & \cellcolor{green!45}$^\downarrow$0.0424
 ± 0.0102
 & 82.75 ± 0.92 & \cellcolor{green!10}0.4551
± 0.1137
 & \cellcolor{green!45}$^\downarrow$0.0823
 ± 0.0164
 \\ 
\hline
\multicolumn{5}{c}{\textbf{Monte Carlo Dropout}}        
        \\ \hline Abt-Buy & \cellcolor{green!10}0.0193
 ± 0.0016
 & \cellcolor{red!45}$^\downarrow$90.68
 ± 0.92
 & \cellcolor{red!10}0.9504
 ± 0.0298
 & \cellcolor{red!10}0.0574
 ± 0.0037
 \\ 
        DBLP-ACM-S & \cellcolor{green!10}0.0038
 ± 0.0010
 & \cellcolor{green!10}98.83
 ± 0.32
 & \cellcolor{red!10}0.8716
 ± 0.1538
 & \cellcolor{red!10}0.0333
 ± 0.0096
 \\ 
        DBLP-ACM-D & \cellcolor{green!10}0.0042
 ± 0.0011
 & \cellcolor{green!10}98.90
 ± 0.21
 & \cellcolor{red!10}0.7207
 ± 0.1148
 & \cellcolor{green!10}0.0286
 ± 0.0096
 \\ 
        iTunes-Amazon-S & \cellcolor{green!10}0.0381

 ± 0.0084

 & \cellcolor{green!10}90.87

 ± 1.37

 & \cellcolor{green!10}0.3008
 ± 0.1470
 & \cellcolor{green!10}0.0495
 ± 0.0096
 \\ 
        iTunes-Amazon-D & \cellcolor{green!45}$^\downarrow$0.0381
 ± 0.0124
 & 91.50
 ± 1.90
 & \cellcolor{red!10}0.4036
 ± 0.3180
 & \cellcolor{red!10}0.0718
 ± 0.0235
\\ 
        Company & \cellcolor{green!10}0.0543
 ± 0.0085
 & \cellcolor{green!10}82.75
 ± 0.86
 & \cellcolor{green!10}0.5137
 ± 0.0928
 & \cellcolor{green!10}0.0946
 ± 0.0156
 \\ 
\hline
\multicolumn{5}{c}{\textbf{Ensembles}}        \\ \hline Abt-Buy & \cellcolor{green!45}$^\downarrow$0.0173

 ± 0.0005

 & \cellcolor{red!10}90.78

 ± 0.34

 & \cellcolor{green!45}$^\downarrow$0.8669

 ± 0.0316

 & \cellcolor{red!45}$^\uparrow$0.0672

 ± 0.0031

 \\ 
        DBLP-ACM-S & \cellcolor{red!10}0.0057

 ± 0.0023

 & \cellcolor{green!10}98.89

 ± 0.20

 & \cellcolor{red!10}0.7914

 ± 0.2040
 & \cellcolor{red!10}0.0370
 ± 0.0096

 \\ 
        DBLP-ACM-D & \cellcolor{red!10}0.0052

 ± 0.0007

 & \cellcolor{red!45}$^\downarrow$98.51

 ± 0.15

 & \cellcolor{red!45}$^\uparrow$0.8557

 ± 0.1063

 & \cellcolor{red!45}$^\uparrow$0.0439

 ± 0.0026

 \\ 
        iTunes-Amazon-S & \cellcolor{green!45}$^\downarrow$0.0333

 ± 0.0022

 & \cellcolor{green!10}91.61

 ± 0.95

 & \cellcolor{red!45}$^\uparrow$0.6869

 ± 0.1421

 & \cellcolor{red!45}$^\uparrow$0.0948

 ± 0.0176

 \\ 
        iTunes-Amazon-D & \cellcolor{green!10}0.0438

 ± 0.0123

 & \cellcolor{red!10}91.34

 ± 2.52

 & \cellcolor{red!45}$^\uparrow$0.5904

 ± 0.0296

 &\cellcolor{red!45}$^\uparrow$0.0950

 ± 0.0143

\\ 
        Company & *
 & *
 & *
 & *
\\
\end{tabular}
\caption{The mean ECE, $F_1$ score, MCE, and RMSCE results over five runs, for the confidence calibration methods and for the baseline predictions, on all datasets, along with standard deviations. $F_1$ scores are reported to two decimal places. The other metrics are reported to four decimal places. Green cells signify that a result is better compared to the result for the uncalibrated pipeline; red cells signify that a result is worse compared to the result for the uncalibrated pipeline. Saturated colours indicate that the performance difference is significant  ($\alpha=0.05$), with arrows showing if the difference is negative or positive. *: Company dataset results were not gathered for the Ensembles method due to computational constraints.}
\label{tab:calibration_results}
\end{table*}

\begin{figure*}[]
     \centering
     \begin{subfigure}[b]{0.45\textwidth}
         \centering
         \includegraphics[width=\textwidth]{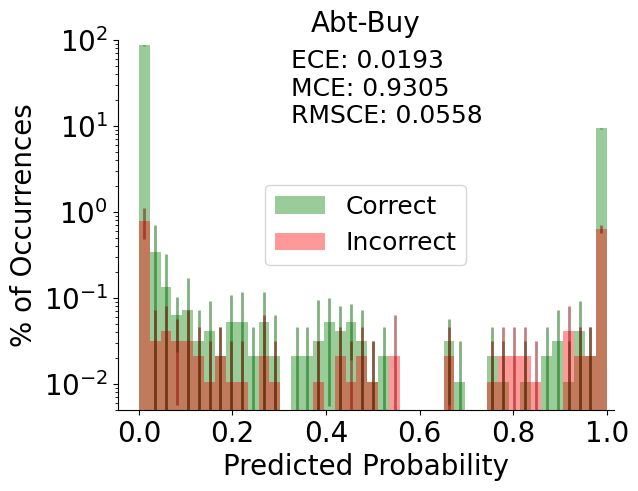}
         \caption{}
         \label{fig:double_confidence_histograms_abt_buy}
     \end{subfigure}
     \hfill
     \begin{subfigure}[b]{0.45\textwidth}
         \centering
         \includegraphics[width=\textwidth]{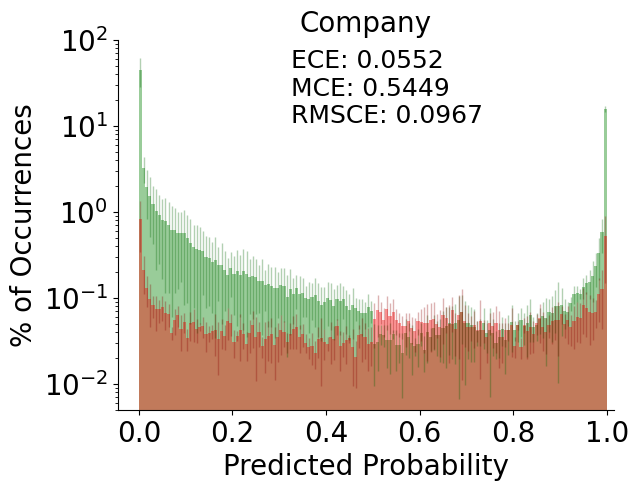}
         \caption{}
         \label{fig:double_confidence_histograms_company}
     \end{subfigure} \\
        \caption{The mean confidence histograms over five runs for the Abt-Buy and Company datasets, using the baseline RoBERTa  model predictions, on a logarithmic scale. The distribution of correct prediction values are in green; the distribution of incorrect prediction values are in red. The y-axis presents percentages of occurrences rather than absolute numbers of occurrences. Error bars denote standard deviations. ECE, MCE, and RMSCE values are reported to four decimal places. The same confidence histograms for the other four datasets are presented in Appendix \ref{sec:roberta_confidence_histograms}.}
\label{fig:double_confidence_histograms_results}
\end{figure*}

\subsection{Baseline}

We find that, for all datasets, the RoBERTa EM model produces either very low or very high predicted probabilities, signifying a high overall confidence (Appendix \ref{sec:roberta_confidence_histograms}). High confidence outputs do not necessarily signify miscalibration. The high baseline $F_1$ scores in Table \ref{tab:calibration_results}, especially for the DBLP-ACM datasets, suggest that the model makes few errors and can justifiably be confident in its predictions. Still, however, we observe that the model produces very high confidence levels even for the datasets where the classification $F_1$ scores are around 90 or lower. Confidence histograms that separately display the distributions of correct and incorrect predictions for the datasets also suggest a miscalibration. Two of these confidence histograms are presented as examples in Figure \ref{fig:double_confidence_histograms_results}. For a well-calibrated pipeline, there should be minimal overlap between the distributions of correct and incorrect predictions in such histograms. Figure \ref{fig:double_confidence_histograms_results} shows that this is not the case.

As visible in Table \ref{tab:calibration_results}, the baseline ECEs are lowest for the DBLP-ACM datasets. These are also the datasets for which the baseline RoBERTa model achieves the highest $F_1$ scores. The ECEs are highest for the iTunes-Amazon, and Company datasets. While the Company datasets' ECEs may in part be due to their challenging EM data, this does not explain the iTunes-Amazon ECEs.

Since the ECE is a measure that is weighted by the number of data points, it is most influenced by the extreme prediction probabilities. After all, these occur most often. The RMSCE is, compared to the ECE, influenced more by large errors between the predicted probability and the empirical probability. The reported values for this RMSCE metric are consistently higher than the reported ECEs. This is especially the case for the DBLP-ACM, Company, and Abt-Buy datasets. The reliability diagrams presented in Appendix \ref{sec:roberta_reliability_diagrams} present an explanation for the higher RMSCEs, showing that there exist large errors between the predicted probabilities and the empirical probabilities for all datasets. 

The MCE measures the maximum discrepancy between predicted and empirical probabilities. Figure \ref{fig:calibration_curves} in Appendix \ref{sec:roberta_reliability_diagrams} shows that this difference is large for most datasets, resulting in high MCEs. However, these maximum discrepancies occur for predicted probabilities with few data points, as the figures in Appendix \ref{sec:roberta_confidence_histograms} show.

We find no correlation between the ECE, MCE, or RMSCE metric values and the datasets' $F_1$ scores, sizes, or mean entry pair sizes.

\subsection{Calibration Methods}

\subsubsection{Temperature Scaling} As Table \ref{tab:calibration_results} shows, for the Temperature Scaling method, the ECE significantly decreases for the Abt-Buy, DBLP-ACM-Structured, iTunes-Amazon-Structured, and Company datasets when compared to the baseline. For the other datasets, the ECE decreases, but not significantly. The percentage decrease in ECE compared to the baseline results across the public datasets ranges from 8.05\% (for iTunes-Amazon-Dirty) to 23.83\% (for Abt-Buy).

For the majority of datasets, however, the changes in MCE and RMSCE are not significant. This is likely because the temperature parameter that is used for Temperature Scaling is optimised using the ECE, and not the MCE or RMSCE. We therefore suggest, for practical applications, to consider whether to prioritize reducing the mean error, larger errors, or the maximum error in calibration. The temperature parameter can then be optimised on respectively the ECE, RMSCE, or MCE.

Figure \ref{fig:validation_set_performance_temperature_scaling} in Appendix \ref{sec:appendix_temperature_scaling} shows that for every dataset and run, there seems to be a clear optimum in the temperature parameter value when optimising on the validation set. As shown in Table \ref{tab:experiment_3_temperatures}, the optimal temperature values are typically greater than 1.00. This means that the resulting sigmoid scores are drawn closer to 0.5 compared to when no temperature scaling is applied. This further demonstrates that the baseline probability predictions of the RoBERTa EM model tend to be overconfident.

\subsubsection{Monte Carlo Dropout} For Monte Carlo Dropout, the ECE often decreases compared to the baseline, though this difference is almost always not significant. For Abt-Buy, Monte Carlo Dropout leads to a significant decrease in the $F_1$ score.  

Figure \ref{fig:validation_set_performance_dropout_eces}
in Appendix \ref{sec:appendix_mc_dropout} shows that for none of the trained models and datasets, there seems to be a very clear optimal dropout probability parameter value when optimising on the validation set. Only very high dropout values negatively impact the ECE. %
The same pattern is observed in Figure \ref{fig:validation_set_performance_dropout_f1_scores}
of the Appendix \ref{sec:appendix_mc_dropout}. This figure also suggests that a considerable dropout probability can be used on most datasets without weakening the performance. Table \ref{tab:experiment_3_dropout_probabilities} further demonstrates this, as for most datasets, the optimal dropout probability lies between 0.5 and 1.0. For two datasets, the optimal dropout probabilities are even above 0.8. Table \ref{tab:experiment_3_dropout_probabilities} moreover shows that the mean optimal dropout probabilities and standard deviations can vary considerably among datasets, suggesting a lack of generalisability for the dropout parameter. On the other hand, again, Figure \ref{fig:validation_set_performance_dropout_eces}
shows that there are no clear optima of the dropout probabilities per dataset on the validation ECEs.

Monte Carlo Dropout causes no significant changes in MCE or RMSCE. Like for Temperature Scaling, we suggest to optimise on the ECE, RMSCE, or MCE depending on the desired confidence calibration behaviour.

\subsubsection{Ensembles} For the Ensembles calibration method, the ECE decreases for the Abt-Buy and iTunes-Amazon-Structured datasets. For the DBLP-ACM and iTunes-Amazon-Dirty datasets, the change is not significant. With regard to the $F_1$ score, the results are also often not significant, although the $F_1$ score for the DBLP-ACM-Dirty dataset does decrease significantly.

Monte Carlo performs multiple sub-runs with dropout during inference. Ensembles train multiple models using differently initialised weights. For both methods, the predictions of respectively these sub-runs and models are averaged and used as final prediction probabilities. A possible reason for the limited significant improvements in ECEs for the Monte Carlo Dropout and Ensemble methods is the similarity in the predictions of the sub-runs and models. After all, the only difference in producing these predictions is, for Monte Carlo Dropout, the dropout in the final classification layer, or, for Ensembles, the initialisation of this classification layer. The inputs to this classification layer come from the same pre-trained model checkpoint, resulting in highly correlated sub-run or model predictions. This strong correlation likely limits the effectiveness of both Monte Carlo Dropout and Ensembles. Xiao and colleagues also describe this drawback for Ensembles \cite{xiao_uncertainty_2022}.

%% file: sections/conclusion.tex
We compare the confidence calibration of baseline RoBERTa probability predictions without any use of confidence calibration methods, to the confidence calibration using Temperature Scaling, Monte Carlo Dropout and Ensembles as confidence calibration methods for EM. 

We find that the ECE performance and overall confidence calibration performance for RoBERTa's performance on EM, without using any confidence calibration methods, is reasonable, but often overconfident, with ECE scores ranging from 0.0043 to 0.0552, leaving room for improvement.

We find Temperature Scaling to work best, compared to Monte Carlo Dropout and Ensembles, in improving a RoBERTa model's ECEs for EM, reducing ECE scores by up to 23.83\%. This is a simple method that can easily be implemented in practical settings.

We find that neither Temperature Scaling, Monte Carlo Dropout, nor Ensembles have consistently significant effects on the $F_1$ scores of the the RoBERTa EM model.

%% file: sections/future_work.tex
Overall, the ECEs reported for the baseline RoBERTa EM model results are slightly higher than those reported for RoBERTa by Desai and Durrett \cite{desai_calibration_2020} and slightly lower to those reported for RoBERTa by Xiao and colleagues \cite{xiao_uncertainty_2022}. Both studies applied the model to natural language processing tasks other than EM. It would be interesting for future research to investigate the cause of these differences in metric values. 

Another avenue for future research is to combine confidence calibration methods for EM. For example, \citet{rahaman_uncertainty_2021} found that using Ensembles, and applying Temperature Scaling to the averaged sigmoid scores can reduce ECE scores by half compared to just using Ensembles, on image classification tasks. Temperature Scaling could be combined with Monte Carlo Dropout in the same way. 

Additionally, future work could leverage the individual variances in the sigmoid scores produced by Monte Carlo Dropout and Ensembles. If these variances are high, the confidence levels can be lowered accordingly, potentially improving calibration. By incorporating variance-based adjustments, it might be possible to create more reliable confidence estimates and further enhance the overall performance of the RoBERTa pipeline. Additionally, entry pairs with large variances in their sigmoid scores can be more closely analyzed to gain deeper insights into the pipeline's prediction patterns.

%% file: sections/limitations.tex
Recent years have seen massive advances in LLMs, yet this study focuses on a relatively small-scale model compared to state-of-the-art architectures. The academic community has extensively researched derivatives of the BERT model, and smaller models remain practical for deployment on limited computational resources facilities. However, an important next step is to extend these model calibration experiments to larger models and evaluate their trustworthiness capabilities under similar conditions.

It is worth noting that the ECE, MCE, and RMSCE metrics are not without limitations in accurately capturing confidence calibration. To illustrate this, suppose there is an EM dataset with 50\% `match' labels and 50\% `no-match' labels. If a model would only output predicted probabilities of 0.5, the ECE, MCE and RMSCE would all be zero, suggesting approximately perfect calibration. Yet, the model's predicted probabilities would be entirely uninformative.

%% file: sections/ethics_statement.tex
We recognize that while LLMs have proven to be successful in EM tasks, these models also pose risks. An example of this is the potential for bias in LLM outputs, discussed in detail in the paper ``On the Dangers of Stochastic Parrots'' by \citet{bender_dangers_2021}. Since models such as RoBERTa are pre-trained on large amounts of data that reflect societal biases, these prejudices can be incorporated into and potentially be amplified in EM predictions. Moreover, LLMs operate as black-box models, providing little transparency on their decision-making processes. In this research, we explored this problem through a study on confidence calibration, so that it can be mitigated. Enhancing transparency can help avoid incorrect downstream decisions and make it easier to analyze and rectify erroneous or misleading outputs.

%% file: sections/appendix.tex
\section{RoBERTa EM Performance}
\label{sec:roberta_em_performance}

Table \ref{tab:roberta_public_datasets_results} presents the mean $F_1$ score, precision, recall and inference time for the baseline RoBERTa model.

\begin{table*}[ht]
    \centering
    \begin{tabular}{l|llll}
    \hline
        \textbf{Dataset} & $\mathbf{F_1}$ & \textbf{Precision} & \textbf{Recall} & \textbf{Inference time (ms)} \\ \hline
        Abt-Buy & 90.81 ± 0.85 & 91.86 ± 0.55 & 89.81 ± 1.82 & 1.43 ± 0.01 \\ 
        DBLP-ACM-Structured & 98.78 ± 0.40 & 98.83 ± 0.73 & 98.74 ± 0.12 & 2.04 ± 0.01 \\ 
        DBLP-ACM-Dirty & 98.85 ±  0.18 & 98.88 ± 0.50 & 98.83 ± 0.25 & 2.06 ± 0.01 \\
        iTunes-Amazon-Structured & 90.53 ± 1.64 & 93.22 ± 4.80 & 88.15 ± 1.66 & 0.32 ± 0.05 \\ 
        iTunes-Amazon-Dirty & 91.50 ± 1.90 & 87.81 ± 2.83 & 95.56 ± 1.66 & 0.28 ± 0.07 \\ 
        Company & 82.75 ± 0.92 & 82.20 ± 2.95 & 83.40 ± 1.53 & 2.51 ± 0.00 \\ 
    \end{tabular}
    \caption{The mean $F_1$ score, precision, recall, and inference time (in milliseconds) for the RoBERTa EM model for all datasets, along with the standard deviations. Metrics are taken over five randomly initialised runs and reported to two decimal places.}
\label{tab:roberta_public_datasets_results}
\vspace{-4mm}
\end{table*}

\section{RoBERTa Confidence Histograms}
\label{sec:roberta_confidence_histograms}

\begin{figure*}[]
     \centering
     \begin{subfigure}[b]{0.45\textwidth}
         \centering
         \includegraphics[width=\textwidth]{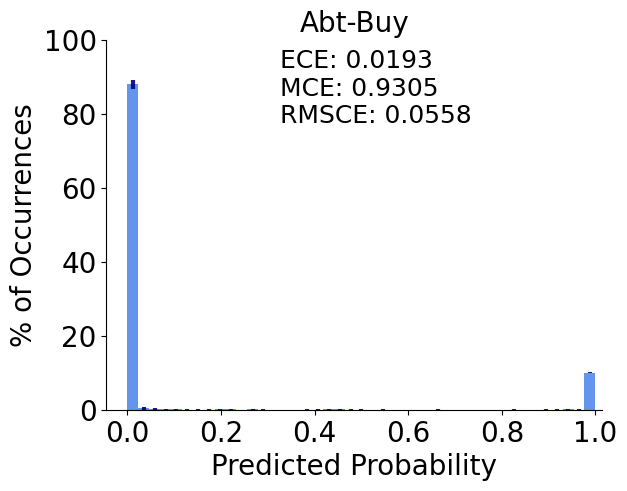}
         \caption{}
         \label{fig:basic_confidence_histograms_abt_buy}
     \end{subfigure}
     \hfill
     \begin{subfigure}[b]{0.45\textwidth}
         \centering
         \includegraphics[width=\textwidth]{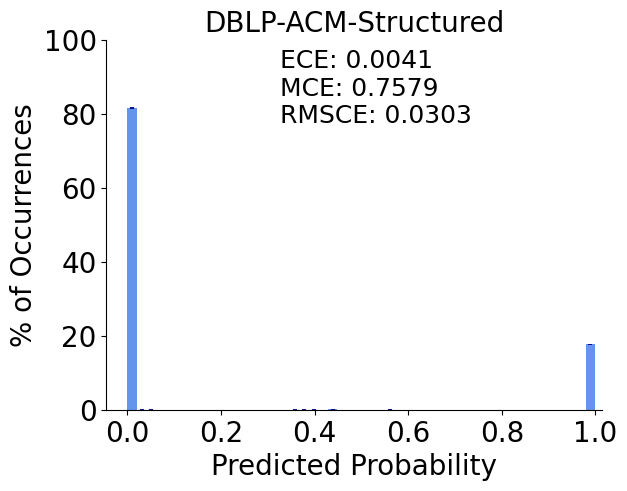}
         \caption{}
         \label{fig:basic_confidence_histograms_dblp_acm_structured}
     \end{subfigure} 
     \\
      \begin{subfigure}[b]{0.45\textwidth}
         \centering
         \includegraphics[width=\textwidth]{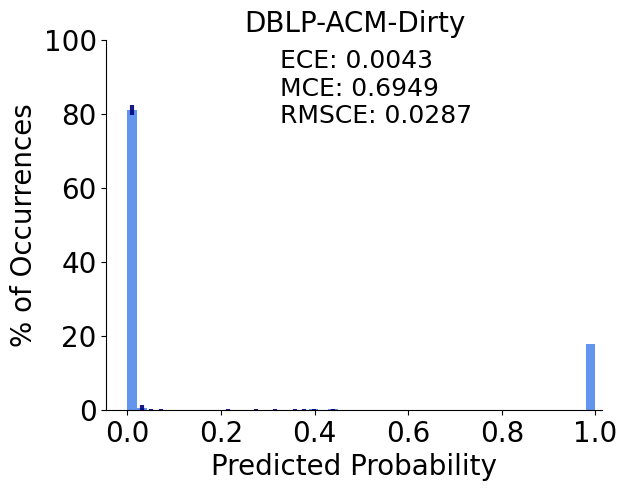}
         \caption{}
         \label{fig:basic_confidence_histograms_dblp_acm_dirty}
     \end{subfigure}
     \hfill
          \begin{subfigure}[b]{0.45\textwidth}
         \centering
         \includegraphics[width=\textwidth]{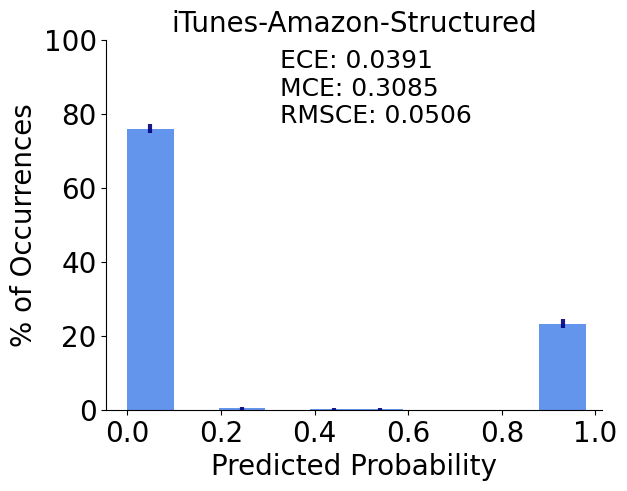}
         \caption{}
         \label{fig:basic_confidence_histograms_itunes_amazon_structured}
     \end{subfigure} 
     \\ 
           \begin{subfigure}[b]{0.45\textwidth}
         \centering
         \includegraphics[width=\textwidth]{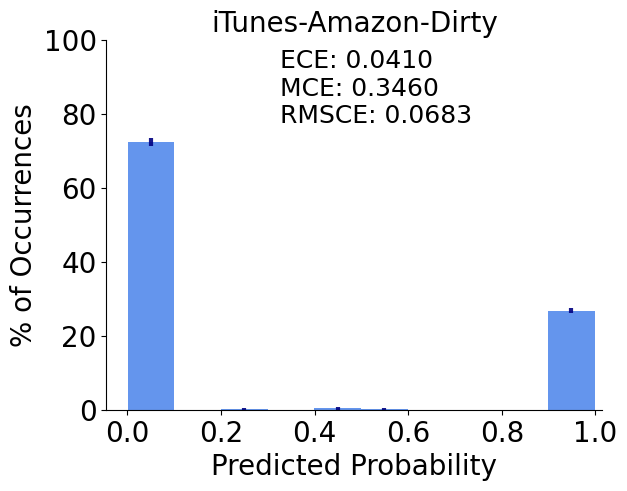}
         \caption{}
         \label{fig:basic_confidence_histograms_itunes_amazon_dirty}
     \end{subfigure}
     \hfill
          \begin{subfigure}[b]{0.45\textwidth}
         \centering
         \includegraphics[width=\textwidth]{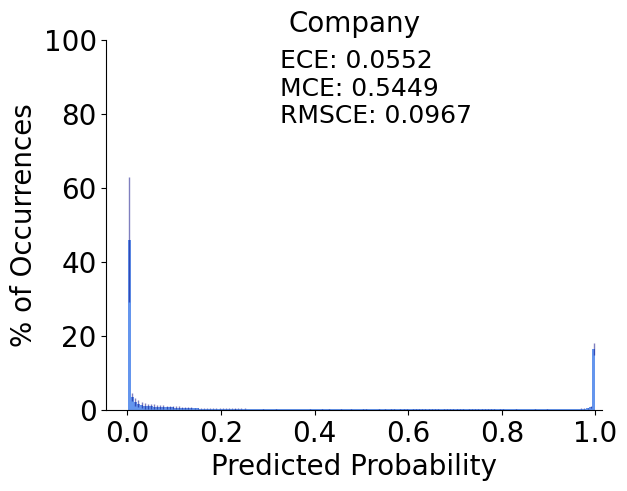}
         \caption{}
         \label{fig:basic_confidence_histograms_company}
     \end{subfigure} \\ 
        \caption{The mean confidence histograms over five runs for all datasets, using the baseline RoBERTa model predicted probabilities. The y-axis presents percentages of occurrences rather than absolute numbers of occurrences. Error bars denote standard deviations. ECE, MCE, and RMSCE values are reported to four decimal places.}
        \label{fig:basic_confidence_histograms}
\end{figure*}

The confidence histograms for all datasets, using the baseline RoBERTa model predicted probabilities and a number of bins = $\sqrt{|\mathcal{D}|}$, are presented in Figure \ref{fig:basic_confidence_histograms}.

\begin{figure*}[]
     \centering
     \begin{subfigure}[b]{0.45\textwidth}
         \centering
         \includegraphics[width=\textwidth]{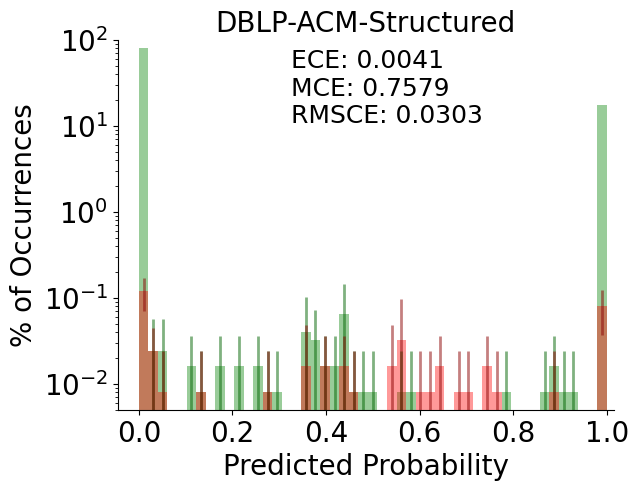}
         \caption{}
         \label{fig:double_confidence_histograms_dblp_acm_structured}
     \end{subfigure} 
     \hfill
      \begin{subfigure}[b]{0.45\textwidth}
         \centering
         \includegraphics[width=\textwidth]{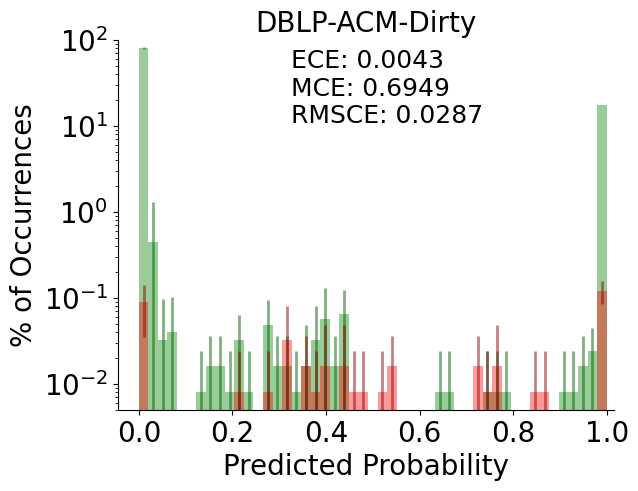}
         \caption{}
         \label{fig:double_confidence_histograms_dblp_acm_dirty}
     \end{subfigure}
     \\
          \begin{subfigure}[b]{0.45\textwidth}
         \centering
         \includegraphics[width=\textwidth]{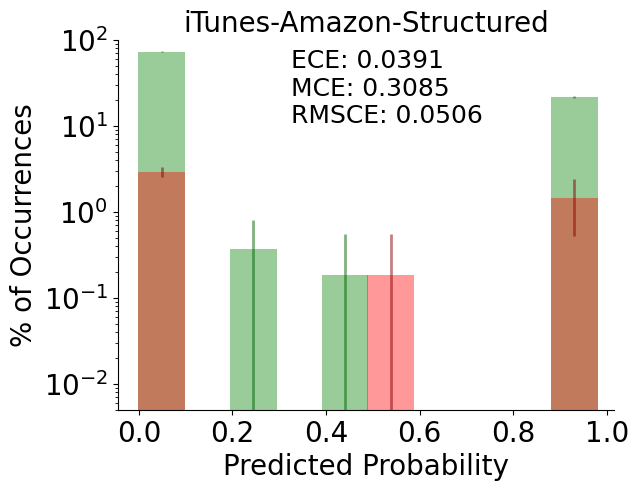}
         \caption{}
         \label{fig:double_confidence_histograms_itunes_amazon_structured}
     \end{subfigure} 
     \hfill
           \begin{subfigure}[b]{0.45\textwidth}
         \centering
         \includegraphics[width=\textwidth]{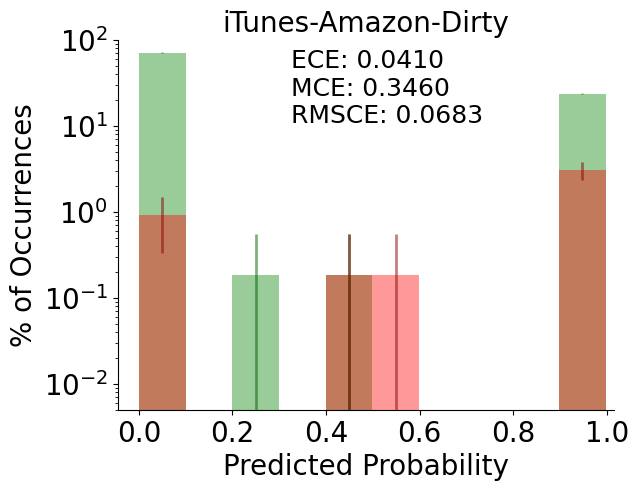}
         \caption{}
         \label{fig:double_confidence_histograms_itunes_amazon_dirty}
     \end{subfigure}
     \\
        \caption{The mean confidence histograms over five runs for the DBLP-ACM-Structured, DBLP-ACM-Dirty, iTunes-Amazon-Structured and iTunes-Amazon-Dirty datasets, using the baseline RoBERTa model predictions, on a logarithmic scale. The distribution of correct prediction values are in green; the distribution of incorrect prediction values are in red. The y-axis presents percentages of occurrences rather than absolute numbers of occurrences. Error bars denote standard deviations. ECE, MCE, and RMSCE values are reported to four decimal places.}
\label{fig:double_confidence_histograms}
\end{figure*}

Figure \ref{fig:double_confidence_histograms} shows confidence histograms that are similar to those in Figure \ref{fig:basic_confidence_histograms}. Histograms are presented for all datasets, using the baseline RoBERTa model predicted probabilities and a number of bins = $\sqrt{|\mathcal{D}|}$. For Figure \ref{fig:double_confidence_histograms}, correct and incorrect predictions are plotted individually. Moreover, the distribution of predicted values is plotted on a logarithmic scale, so that smaller effects are easier to see. Confidence histograms for four out of the six datasets are shown. The confidence histograms for the Abt-Buy and Company datasets are presented in Section \ref{sec:results}.

\section{RoBERTa Reliability Diagrams}
\label{sec:roberta_reliability_diagrams}

\begin{figure*}[]
     \centering
     \begin{subfigure}[b]{0.45\textwidth}
         \centering
         \includegraphics[width=\textwidth]{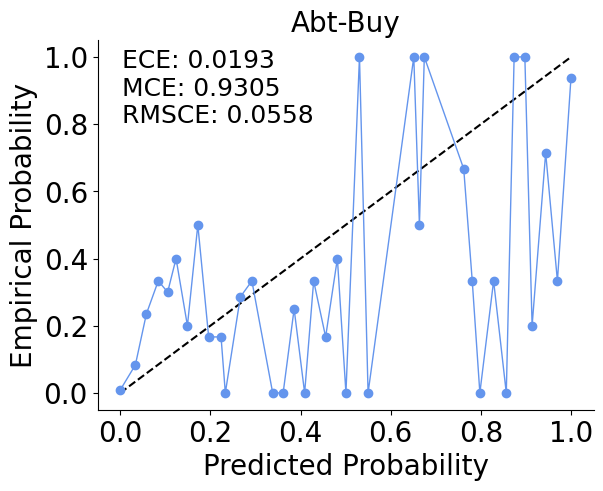}
         \caption{}
         \label{fig:calibration_curve_abt_buy}
     \end{subfigure}
     \hfill
     \begin{subfigure}[b]{0.45\textwidth}
         \centering
         \includegraphics[width=\textwidth]{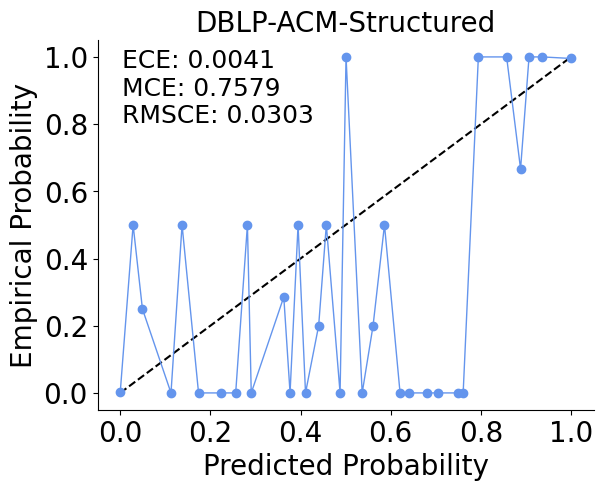}
         \caption{}
         \label{fig:calibration_curve_dblp_acm_structured}
     \end{subfigure} 
     \\
      \begin{subfigure}[b]{0.45\textwidth}
         \centering
         \includegraphics[width=\textwidth]{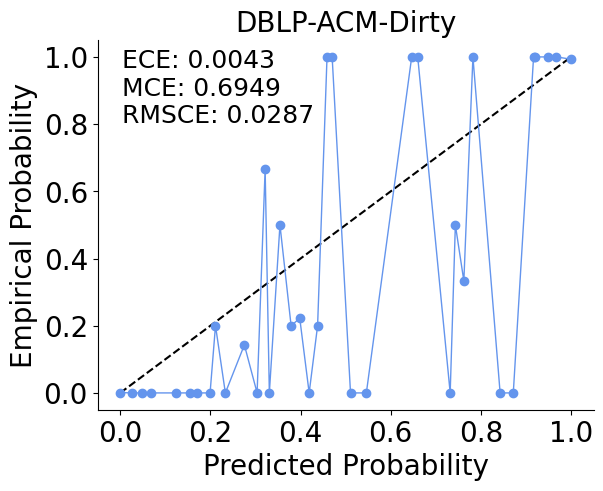}
         \caption{}
         \label{fig:calibration_curve_dblp_acm_dirty}
     \end{subfigure}
     \hfill
          \begin{subfigure}[b]{0.45\textwidth}
         \centering
         \includegraphics[width=\textwidth]{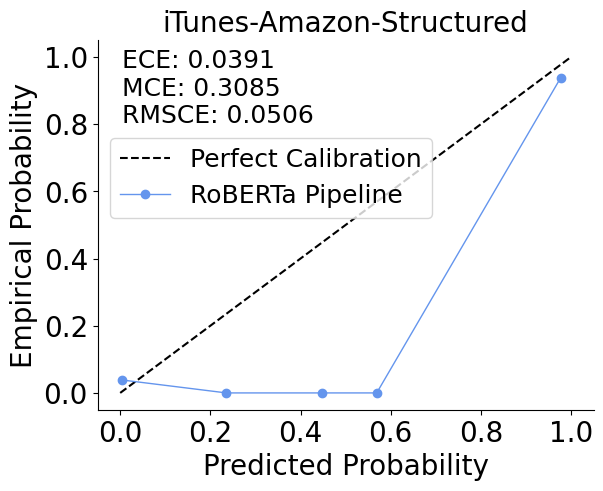}
         \caption{}
         \label{fig:calibration_curve_itunes_amazon_structured}
     \end{subfigure} 
     \\   
           \begin{subfigure}[b]{0.45\textwidth}
         \centering
         \includegraphics[width=\textwidth]{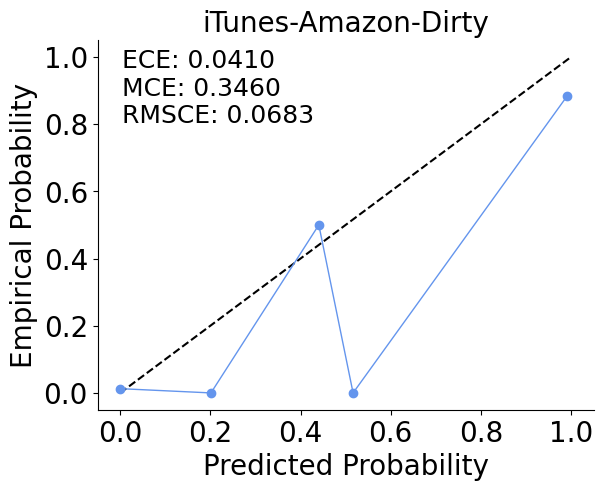}
         \caption{}
         \label{fig:calibration_curve_itunes_amazon_dirty}
     \end{subfigure}
     \hfill
          \begin{subfigure}[b]{0.45\textwidth}
         \centering
         \includegraphics[width=\textwidth]{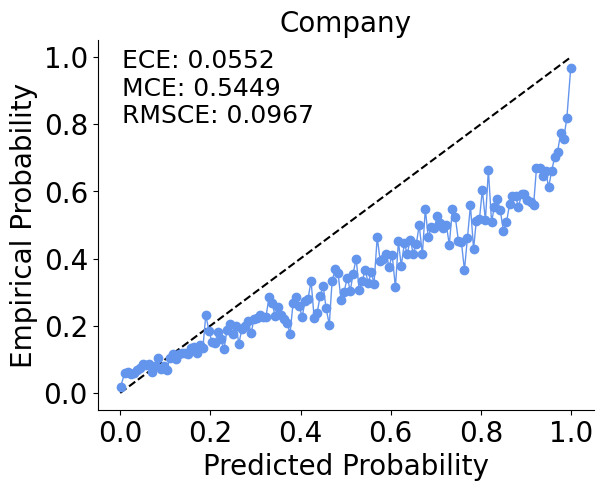}
         \caption{}
         \label{fig:calibration_curve_company}
     \end{subfigure} \\
        \caption{The reliability diagrams using data from five runs for all datasets, using the baseline RoBERTA model predictions. ECE, MCE, and RMSCE values are reported to four decimal digits. Note that for some of the datasets, data is missing for certain predicted probability bins. This is because there were no predictions found within that bin. A diagonal is plotted to represent approximately perfect calibration.}
\label{fig:calibration_curves}
\end{figure*}

Figure \ref{fig:calibration_curves} presents the mean reliability diagrams for all datasets, using the baseline RoBERTa model probability predictions and a number of bins, or dots, = $\sqrt{|\mathcal{D}|}$. When a dot is missing, this means that there are no predictions within that predicted probability bin. A diagonal line representing approximately perfect calibration is plotted as well. 

\section{Detailed Temperature Scaling Results}
\label{sec:appendix_temperature_scaling}

\begin{figure*}[h]
     \centering
     \begin{subfigure}[b]{0.45\textwidth}
         \centering
         \includegraphics[width=\textwidth]{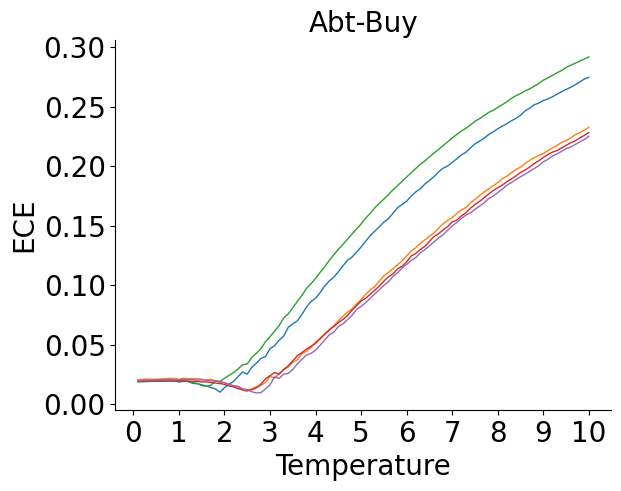}
         \caption{}
         \label{fig:temperatures_eces_abt_buy}
     \end{subfigure}
     \hfill
     \begin{subfigure}[b]{0.45\textwidth}
         \centering
         \includegraphics[width=\textwidth]{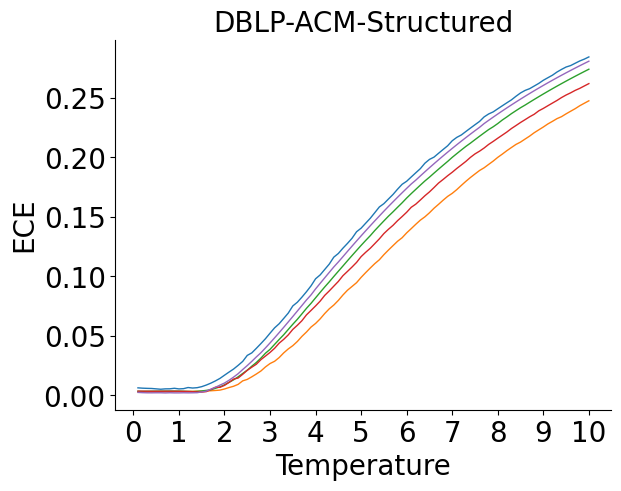}
         \caption{}
         \label{fig:temperatures_eces_dblp_acm_structured}
     \end{subfigure} 
     \\
      \begin{subfigure}[b]{0.45\textwidth}
         \centering
         \includegraphics[width=\textwidth]{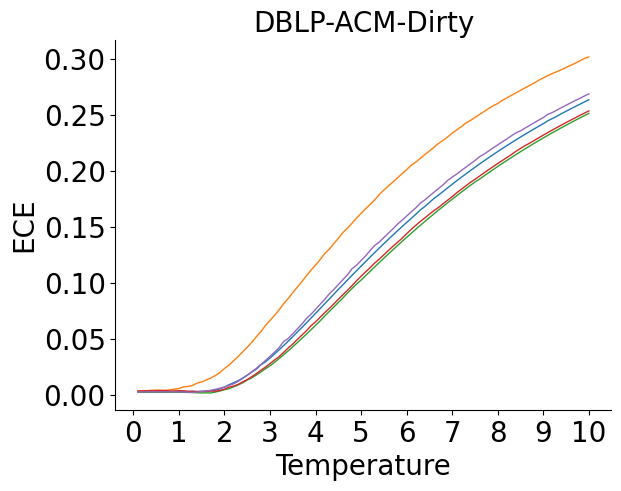}
         \caption{}
         \label{fig:temperatures_eces_dblp_acm_dirty}
     \end{subfigure}
     \hfill
          \begin{subfigure}[b]{0.45\textwidth}
         \centering
         \includegraphics[width=\textwidth]{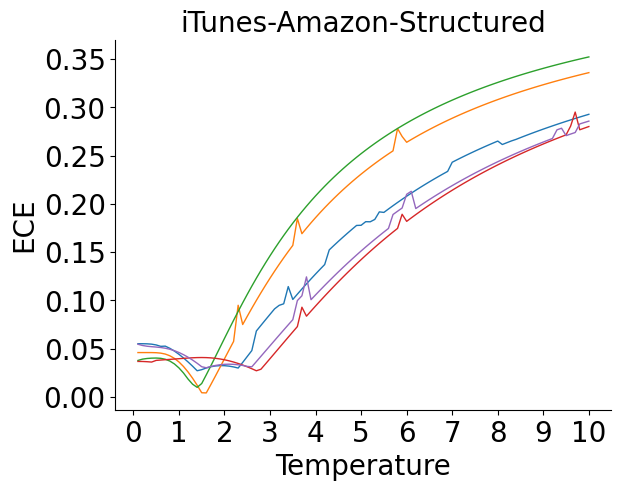}
         \caption{}
         \label{fig:temperatures_eces_itunes_amazon_structured}
     \end{subfigure} 
     \\   
           \begin{subfigure}[b]{0.45\textwidth}
         \centering
         \includegraphics[width=\textwidth]{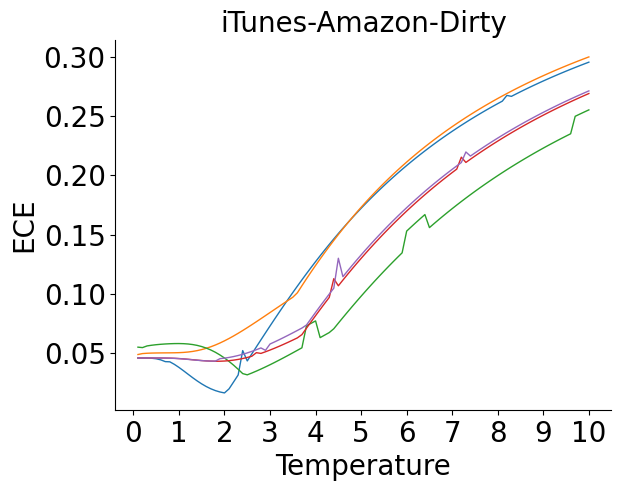}
         \caption{}
         \label{fig:temperatures_eces_itunes_amazon_dirty}
     \end{subfigure}
     \hfill
          \begin{subfigure}[b]{0.45\textwidth}
         \centering
         \includegraphics[width=\textwidth]{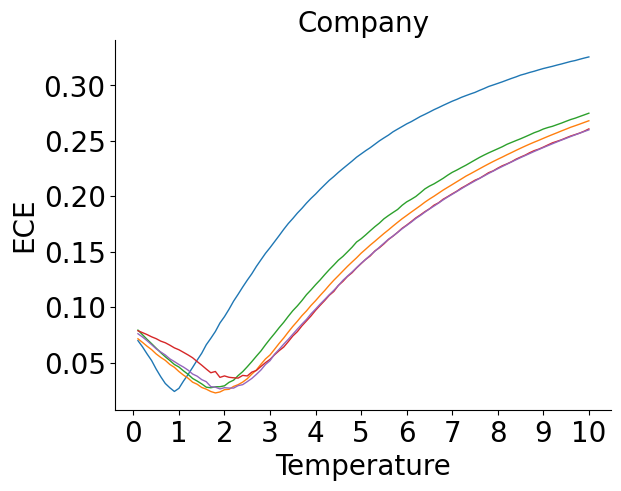}
         \caption{}
         \label{fig:temperatures_eces_company}
     \end{subfigure} \\
        \caption{The effect of the temperature parameter on the ECE for the validation set, for all datasets. Each line denotes one run. Note that the y-axis differs per plot.}
\label{fig:validation_set_performance_temperature_scaling}
\end{figure*}

Figure \ref{fig:validation_set_performance_temperature_scaling}
presents the single parameter gridsearch results for the temperature parameter on the validation sets, for all datasets.

The mean recorded temperature parameter values per dataset are shown in Table \ref{tab:experiment_3_temperatures}.

\begin{table*}
\centering
\footnotesize
\begin{tabular}{lc}
\textbf{Dataset}                              & \textbf{Temperature}  \\ \hline
\multicolumn{1}{l|}{Abt-Buy}                  & 2.24
 ± 0.47\\
\multicolumn{1}{l|}{DBLP-ACM-S}                  & 0.88
 ± 0.50\\
 \multicolumn{1}{l|}{DBLP-ACM-D}                  & 1.00
 ± 0.67\\
 \multicolumn{1}{l|}{iTunes-Amazon-S}                  & 1.74
 ± 0.55\\
 \multicolumn{1}{l|}{iTunes-Amazon-D}                  & 1.64
 ± 0.91\\
 \multicolumn{1}{l|}{Company}                  & 1.72
 ± 0.51\\
\end{tabular}
\caption{The mean temperature parameter value results, taken over five runs, for all datasets, along with the standard deviations. Values are reported to two decimal digits.}
\label{tab:experiment_3_temperatures}
\end{table*}

\section{Detailed Monte Carlo Dropout Results}
\label{sec:appendix_mc_dropout}

\begin{figure*}[]
     \centering
     \begin{subfigure}[b]{0.45\textwidth}
         \centering
         \includegraphics[width=\textwidth]{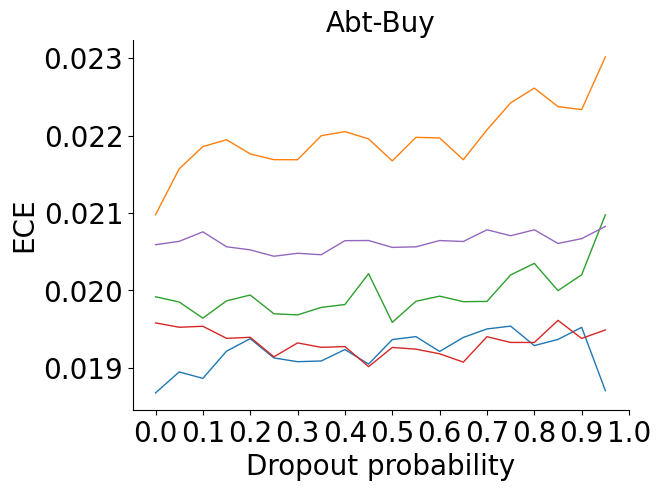}
         \caption{}
         \label{fig:dropout_eces_abt_buy}
     \end{subfigure}
     \hfill
     \begin{subfigure}[b]{0.45\textwidth}
         \centering
         \includegraphics[width=\textwidth]{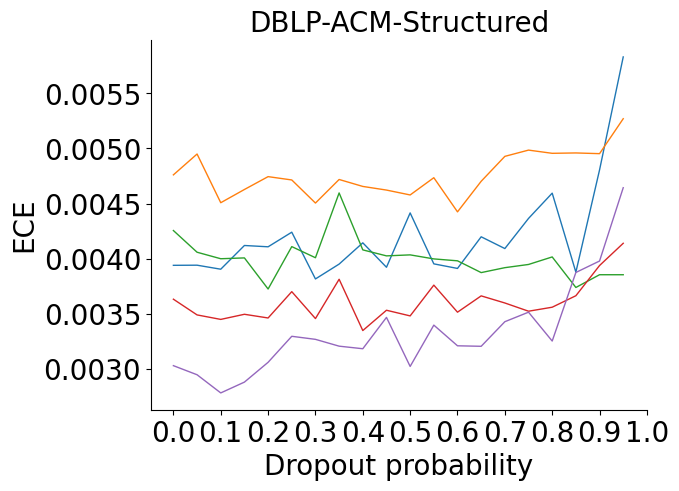}
         \caption{}
         \label{fig:dropout_eces_dblp_acm_structured}
     \end{subfigure} 
     \\
      \begin{subfigure}[b]{0.45\textwidth}
         \centering
         \includegraphics[width=\textwidth]{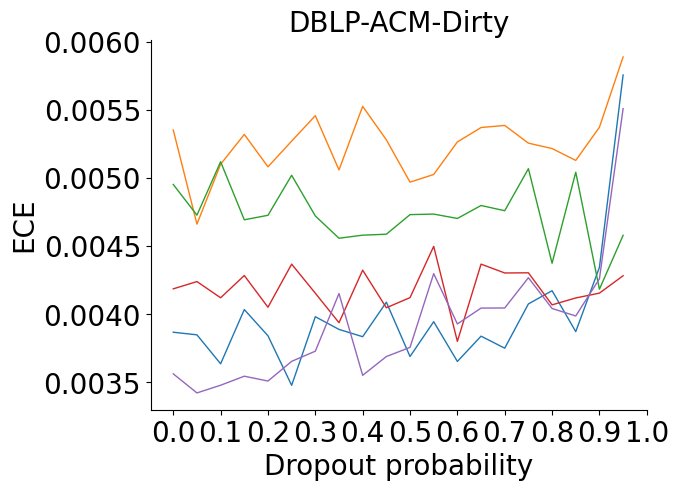}
         \caption{}
         \label{fig:dropout_eces_dblp_acm_dirty}
     \end{subfigure}
     \hfill
          \begin{subfigure}[b]{0.45\textwidth}
         \centering
         \includegraphics[width=\textwidth]{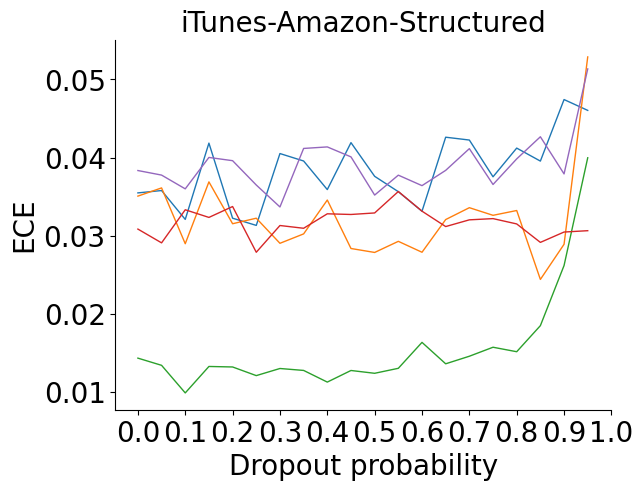}
         \caption{}
         \label{fig:dropout_eces_itunes_amazon_structured}
     \end{subfigure} 
     \\
           \begin{subfigure}[b]{0.45\textwidth}
         \centering
         \includegraphics[width=\textwidth]{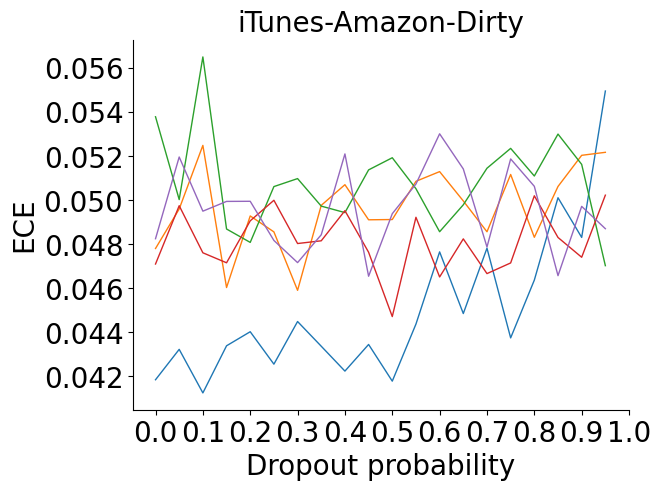}
         \caption{}
         \label{fig:dropout_eces_itunes_amazon_dirty}
     \end{subfigure}
     \hfill
          \begin{subfigure}[b]{0.45\textwidth}
         \centering
         \includegraphics[width=\textwidth]{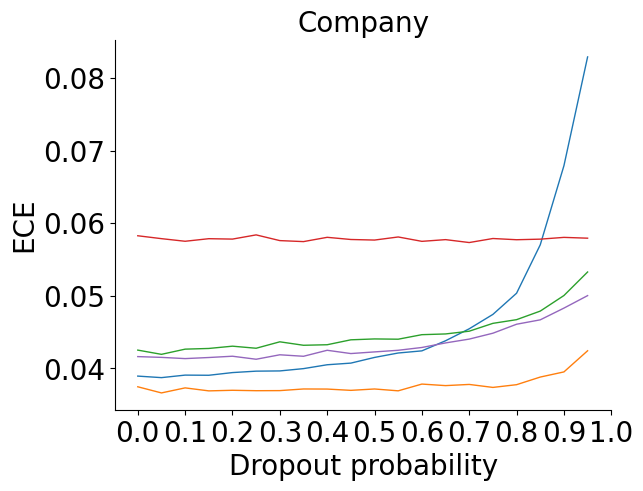}
         \caption{}
         \label{fig:dropout_eces_company}
     \end{subfigure} \\
        \caption{The effect of the dropout probability parameter on the ECE for the validation set, for all datasets. Each line denotes one run. Note that the y-axis differs per plot.}
\label{fig:validation_set_performance_dropout_eces}
\end{figure*}

\begin{figure*}[]
     \centering
     \begin{subfigure}[b]{0.45\textwidth}
         \centering
         \includegraphics[width=\textwidth]{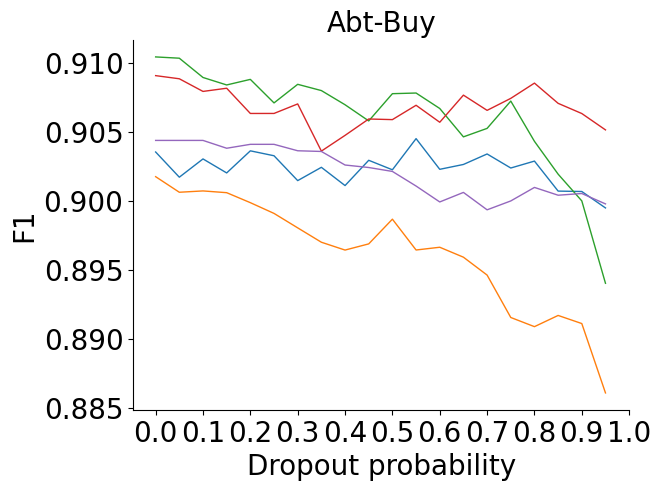}
         \caption{}
         \label{fig:dropout_f1_scores_abt_buy}
     \end{subfigure}
     \hfill
     \begin{subfigure}[b]{0.45\textwidth}
         \centering
         \includegraphics[width=\textwidth]{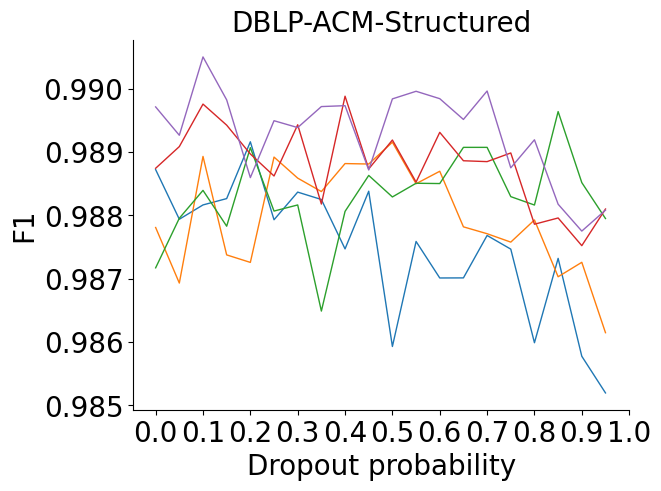}
         \caption{}
         \label{fig:dropout_f1_scores_dblp_acm_structured}
     \end{subfigure} 
     \\
      \begin{subfigure}[b]{0.45\textwidth}
         \centering
         \includegraphics[width=\textwidth]{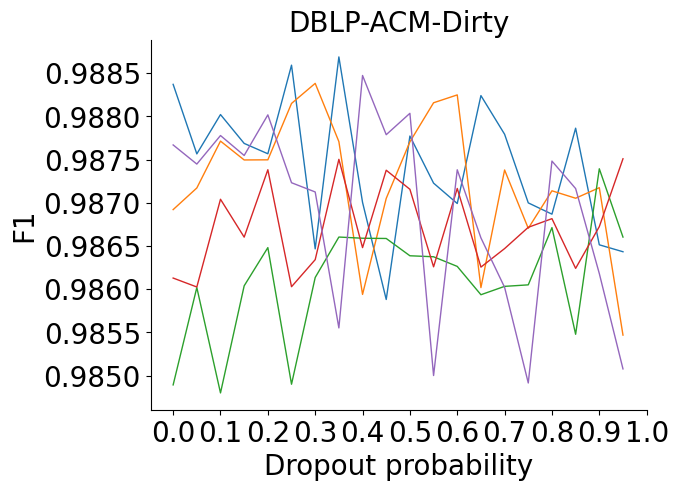}
         \caption{}
         \label{fig:dropout_f1_scores_dblp_acm_dirty}
     \end{subfigure}
     \hfill
          \begin{subfigure}[b]{0.45\textwidth}
         \centering
         \includegraphics[width=\textwidth]{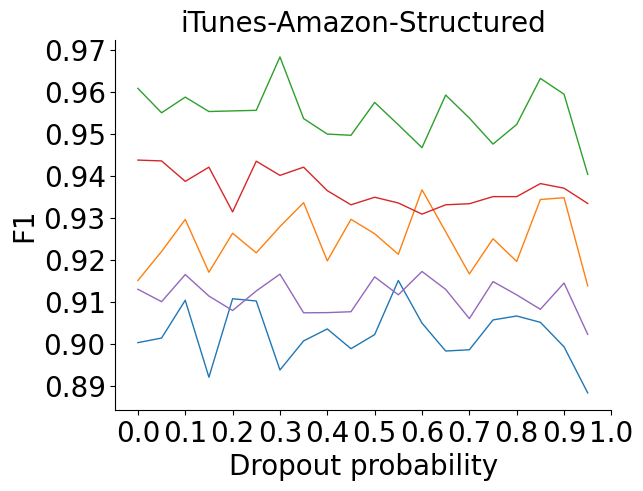}
         \caption{}
         \label{fig:dropout_f1_scores_itunes_amazon_structured}
     \end{subfigure} 
     \\
           \begin{subfigure}[b]{0.45\textwidth}
         \centering
         \includegraphics[width=\textwidth]{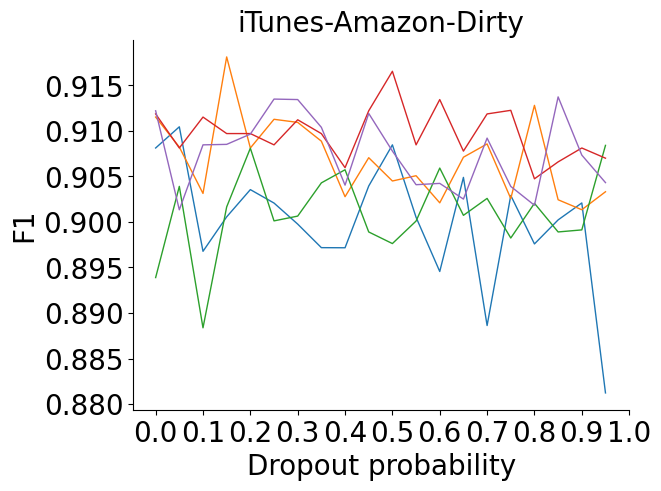}
         \caption{}
         \label{fig:dropout_f1_scores_itunes_amazon_dirty}
     \end{subfigure}
     \hfill
          \begin{subfigure}[b]{0.45\textwidth}
         \centering
         \includegraphics[width=\textwidth]{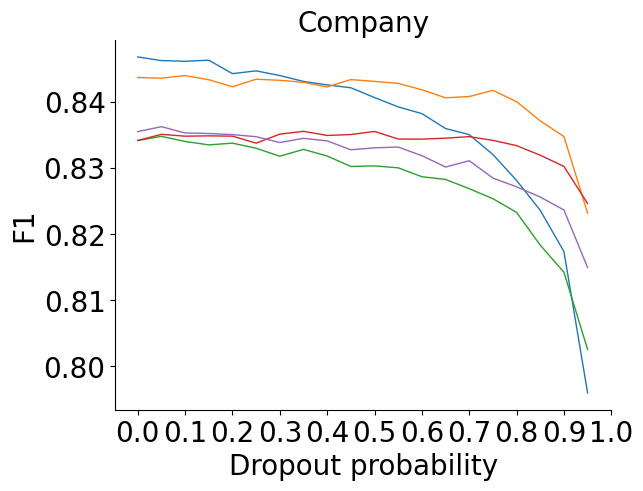}
         \caption{}
         \label{fig:dropout_f1_scores_company}
     \end{subfigure} \\
        \caption{The effect of the dropout probability parameter on the $F_1$ score for the validation set, for all datasets. Each line denotes one run. Note that the y-axis differs per plot.}
\label{fig:validation_set_performance_dropout_f1_scores}
\end{figure*}

Figure \ref{fig:validation_set_performance_dropout_eces} and Figure \ref{fig:validation_set_performance_dropout_f1_scores}
present the single parameter gridsearch results for the dropout parameter on the validation sets, for all datasets. Figure \ref{fig:validation_set_performance_dropout_eces} specifically reports the effect of the dropout probability value on the ECE, while Figure \ref{fig:validation_set_performance_dropout_f1_scores} specifically reports the effect of the dropout probability value on the $F_1$ score.

The mean recorded dropout probability parameter values per dataset are shown in Table \ref{tab:experiment_3_dropout_probabilities}.

\begin{table*}
\centering
\footnotesize
\begin{tabular}{lc}
\textbf{Dataset}                              & \textbf{Dropout probability}  \\ \hline
\multicolumn{1}{l|}{Abt-Buy}                  & 0.39
 ± 0.22\\
\multicolumn{1}{l|}{DBLP-ACM-S}                  & 0.58
 ± 0.40\\
 \multicolumn{1}{l|}{DBLP-ACM-D}                  & 0.56
 ± 0.35\\
 \multicolumn{1}{l|}{iTunes-Amazon-S}                  & 0.85
 ± 0.12\\
 \multicolumn{1}{l|}{iTunes-Amazon-D}                  & 0.91
 ± 0.07\\
 \multicolumn{1}{l|}{Company}                  & 0.50
 ± 0.35\\
\end{tabular}
\caption{The mean dropout probability parameter value results, taken over five runs, for all datasets, using the RoBERTa  model, along with the standard deviations. Values are reported to two decimal digits.}
\label{tab:experiment_3_dropout_probabilities}
\end{table*}

%% file: main.bbl
\begin{thebibliography}{43}
\expandafter\ifx\csname natexlab\endcsname\relax\def\natexlab#1{#1}\fi

\bibitem[{Bender et~al.(2021)Bender, Gebru, McMillan-Major, and
  Shmitchell}]{bender_dangers_2021}
Emily~M Bender, Timnit Gebru, Angelina McMillan-Major, and Shmargaret
  Shmitchell. 2021.
\newblock On the {Dangers} of {Stochastic} {Parrots}: Can {Language} {Models}
  {Be} {Too} {Big}?
\newblock In \emph{Proceedings of the 2021{ACM} {Conference} on {Fairness},
  {Accountability}, and {Transparency}}, pages 610--623.

\bibitem[{Bloothooft et~al.(2015)Bloothooft, Christen, Mandemakers, and
  Schraagen}]{bloothooft_population_2015}
Gerrit Bloothooft, Peter Christen, Kees Mandemakers, and Marijn Schraagen,
  editors. 2015.
\newblock \href {https://doi.org/10.1007/978-3-319-19884-2} {\emph{Population
  {Reconstruction}}}.
\newblock Springer International Publishing AG, Cham.

\bibitem[{Brown et~al.(2020)Brown, Mann, Ryder, Subbiah, Kaplan, Dhariwal,
  Neelakantan, Shyam, Sastry, Askell, Agarwal, Herbert-Voss, Krueger, Henighan,
  Child, Ramesh, Ziegler, Wu, Winter, Hesse, Chen, Sigler, Litwin, Gray, Chess,
  Clark, Berner, McCandlish, Radford, Sutskever, and
  Amodei}]{brown_language_2020}
Tom Brown, Benjamin Mann, Nick Ryder, Melanie Subbiah, Jared~D Kaplan, Prafulla
  Dhariwal, Arvind Neelakantan, Pranav Shyam, Girish Sastry, Amanda Askell,
  Sandhini Agarwal, Ariel Herbert-Voss, Gretchen Krueger, Tom Henighan, Rewon
  Child, Aditya Ramesh, Daniel Ziegler, Jeffrey Wu, Clemens Winter, Chris
  Hesse, Mark Chen, Eric Sigler, Mateusz Litwin, Scott Gray, Benjamin Chess,
  Jack Clark, Christopher Berner, Sam McCandlish, Alec Radford, Ilya Sutskever,
  and Dario Amodei. 2020.
\newblock \href
  {https://papers.nips.cc/paper/2020/hash/1457c0d6bfcb4967418bfb8ac142f64a-Abstract.html}
  {Language {Models} are {Few}-{Shot} {Learners}}.
\newblock In \emph{Advances in {Neural} {Information} {Processing} {Systems}},
  volume~33, pages 1877--1901. Curran Associates, Inc.

\bibitem[{Brunner and Stockinger(2020)}]{brunner_entity_2020}
Ursin Brunner and Kurt Stockinger. 2020.
\newblock Entity {Matching} with {Transformer} {Architectures}-a {Step}
  {Forward} in {Data} {Integration}.
\newblock In \emph{23rd International Conference on Extending Database
  Technology, Copenhagen, 30 March-2 April 2020}, pages 463--473.
  OpenProceedings.

\bibitem[{Chen and Li(2024)}]{chen_calibrating_2024}
Wenlong Chen and Yingzhen Li. 2024.
\newblock \href {https://doi.org/10.48550/arXiv.2303.02444} {Calibrating
  {Transformers} via {Sparse} {Gaussian} {Processes}}.
\newblock ArXiv:2303.02444 [cs, stat].

\bibitem[{Christophides et~al.(2020)Christophides, Efthymiou, Palpanas,
  Papadakis, and Stefanidis}]{christophides_overview_2020}
Vassilis Christophides, Vasilis Efthymiou, Themis Palpanas, George Papadakis,
  and Kostas Stefanidis. 2020.
\newblock \href {https://doi.org/10.1145/3418896} {An {Overview} of
  {End}-to-{End} {Entity} {Resolution} for {Big} {Data}}.
\newblock \emph{ACM Computing Surveys}, 53(6):1--42.

\bibitem[{Dahlin et~al.(2012)Dahlin, Johansson, Kaati, M{\aa}rtenson, and
  Svenson}]{dahlin2012combining}
Johan Dahlin, Fredrik Johansson, Lisa Kaati, Christian M{\aa}rtenson, and
  Pontus Svenson. 2012.
\newblock {Combining} {Entity} {Matching} {Techniques} for {Detecting}
  {Extremist} {Behavior} on {Discussion} {Boards}.
\newblock In \emph{2012 IEEE/ACM International Conference on Advances in Social
  Networks Analysis and Mining}, pages 850--857. IEEE.

\bibitem[{Desai and Durrett(2020)}]{desai_calibration_2020}
Shrey Desai and Greg Durrett. 2020.
\newblock \href {https://doi.org/10.18653/v1/2020.emnlp-main.21} {Calibration
  of {Pre}-trained {Transformers}}.
\newblock In \emph{Proceedings of the 2020 {Conference} on {Empirical}
  {Methods} in {Natural} {Language} {Processing} ({EMNLP})}, pages 295--302,
  Online. Association for Computational Linguistics.

\bibitem[{Devlin et~al.(2019)Devlin, Chang, Lee, and
  Toutanova}]{devlin_bert_2019}
Jacob Devlin, Ming-Wei Chang, Kenton Lee, and Kristina Toutanova. 2019.
\newblock \href {https://doi.org/10.18653/v1/N19-1423} {{BERT}: {Pre}-training
  of {Deep} {Bidirectional} {Transformers} for {Language} {Understanding}}.
\newblock In \emph{Proceedings of the 2019 {Conference} of the {North}
  {American} {Chapter} of the {Association} for {Computational} {Linguistics}:
  {Human} {Language} {Technologies}, {Volume} 1 ({Long} and {Short} {Papers})},
  pages 4171--4186, Minneapolis, Minnesota. Association for Computational
  Linguistics.

\bibitem[{et~al.(2024)}]{openai_gpt-4_2024}
OpenAI et~al. 2024.
\newblock \href {https://doi.org/10.48550/arXiv.2303.08774} {{GPT}-4
  {Technical} {Report}}.
\newblock ArXiv:2303.08774 [cs].

\bibitem[{Gal and Ghahramani(2016)}]{gal_dropout_2016}
Yarin Gal and Zoubin Ghahramani. 2016.
\newblock \href {https://proceedings.mlr.press/v48/gal16.html} {Dropout as a
  {Bayesian} {Approximation}: {Representing} {Model} {Uncertainty} in {Deep}
  {Learning}}.
\newblock In \emph{Proceedings of {The} 33rd {International} {Conference} on
  {Machine} {Learning}}, volume~48, pages 1050--1059. PMLR.
\newblock ISSN: 1938-7228.

\bibitem[{Ghahramani(2015)}]{ghahramani_probabilistic_2015}
Zoubin Ghahramani. 2015.
\newblock \href {https://doi.org/10.1038/nature14541} {Probabilistic {Machine}
  {Learning} and {Artificial} {Intelligence}}.
\newblock \emph{Nature}, 521(7553):452--459.

\bibitem[{Guo et~al.(2017)Guo, Pleiss, Sun, and
  Weinberger}]{guo_calibration_2017}
Chuan Guo, Geoff Pleiss, Yu~Sun, and Kilian~Q. Weinberger. 2017.
\newblock On {Calibration} of {Modern} {Neural} {Networks}.
\newblock In \emph{Proceedings of the 34th {International} {Conference} on
  {Machine} {Learning} - {Volume} 70}, Proceedings of {Machine} {Learning}
  {Research}, pages 1321--1330, Sydney, NSW, Australia. PMLR.

\bibitem[{He et~al.(2016)He, Zhang, Ren, and Sun}]{he_deep_2016}
Kaiming He, Xiangyu Zhang, Shaoqing Ren, and Jian Sun. 2016.
\newblock \href {https://doi.org/10.1109/CVPR.2016.90} {Deep {Residual}
  {Learning} for {Image} {Recognition}}.
\newblock In \emph{2016 {IEEE} {Conference} on {Computer} {Vision} and
  {Pattern} {Recognition} ({CVPR})}, pages 770--778.
\newblock ISSN: 1063-6919.

\bibitem[{Hendrycks and Gimpel(2017)}]{hendrycks2017a}
Dan Hendrycks and Kevin Gimpel. 2017.
\newblock \href {https://openreview.net/forum?id=Hkg4TI9xl} {A {Baseline} for
  {Detecting} {Misclassified} and {Out-of-Distribution} {Examples} in {Neural}
  {Networks}}.
\newblock In \emph{International Conference on Learning Representations}.

\bibitem[{Hinton et~al.(2012)Hinton, Srivastava, Krizhevsky, Sutskever, and
  Salakhutdinov}]{hinton_improving_2012}
Geoffrey~E. Hinton, Nitish Srivastava, Alex Krizhevsky, Ilya Sutskever, and
  Ruslan~R. Salakhutdinov. 2012.
\newblock \href {https://doi.org/10.48550/arXiv.1207.0580} {Improving {Neural}
  {Networks} by {Preventing} {Co}-{Adaptation} of {Feature} {Detectors}}.
\newblock ArXiv:1207.0580 [cs].

\bibitem[{Jaro(1995)}]{jaro_probabilistic_1995}
Matthew~A. Jaro. 1995.
\newblock \href {https://doi.org/10.1002/sim.4780140510} {Probabilistic
  {Linkage} of {Large} {Public} {Health} {Data} {Files}}.
\newblock \emph{Statistics in Medicine}, 14(5-7):491--498.

\bibitem[{Jiang et~al.(2021)Jiang, Araki, Ding, and Neubig}]{jiang_how_2021}
Zhengbao Jiang, Jun Araki, Haibo Ding, and Graham Neubig. 2021.
\newblock \href {https://doi.org/10.1162/tacl_a_00407} {How {Can} {We} {Know}
  {When} {Language} {Models} {Know}? {On} the {Calibration} of {Language}
  {Models} for {Question} {Answering}}.
\newblock \emph{Transactions of the Association for Computational Linguistics},
  9:962--977.

\bibitem[{Konda et~al.(2016)Konda, Das, Suganthan G.~C., Doan, Ardalan,
  Ballard, Li, Panahi, Zhang, Naughton, Prasad, Krishnan, Deep, and
  Raghavendra}]{konda_magellan_2016}
Pradap Konda, Sanjib Das, Paul Suganthan G.~C., AnHai Doan, Adel Ardalan,
  Jeffrey~R. Ballard, Han Li, Fatemah Panahi, Haojun Zhang, Jeff Naughton,
  Shishir Prasad, Ganesh Krishnan, Rohit Deep, and Vijay Raghavendra. 2016.
\newblock \href {https://doi.org/10.14778/2994509.2994535} {Magellan: {Toward}
  {Building} {Entity} {Matching} {Management} {Systems}}.
\newblock \emph{Proceedings of the VLDB Endowment}, 9(12):1197--1208.

\bibitem[{Kumar et~al.(2019)Kumar, Liang, and Ma}]{kumar_verified_2019}
Ananya Kumar, Percy~S Liang, and Tengyu Ma. 2019.
\newblock \href
  {https://papers.nips.cc/paper_files/paper/2019/hash/f8c0c968632845cd133308b1a494967f-Abstract.html}
  {Verified {Uncertainty} {Calibration}}.
\newblock In \emph{Advances in {Neural} {Information} {Processing} {Systems}},
  volume~32. Curran Associates, Inc.

\bibitem[{Köpcke et~al.(2010)Köpcke, Thor, and Rahm}]{kopcke_evaluation_2010}
Hanna Köpcke, Andreas Thor, and Erhard Rahm. 2010.
\newblock \href {https://doi.org/10.14778/1920841.1920904} {Evaluation of
  {Entity} {Resolution} {Approaches} on {Real}-{World} {Match} {Problems}}.
\newblock \emph{Proc. VLDB Endow.}, 3(1-2):484--493.

\bibitem[{Küppers et~al.(2022)Küppers, Haselhoff, Kronenberger, and
  Schneider}]{kuppers_confidence_2022}
Fabian Küppers, Anselm Haselhoff, Jan Kronenberger, and Jonas Schneider. 2022.
\newblock \href {https://doi.org/10.1007/978-3-031-01233-4_8} {Confidence
  {Calibration} for {Object} {Detection} and {Segmentation}}.
\newblock In Tim Fingscheidt, Hanno Gottschalk, and Sebastian Houben, editors,
  \emph{Deep {Neural} {Networks} and {Data} for {Automated} {Driving}:
  {Robustness}, {Uncertainty} {Quantification}, and {Insights} {Towards}
  {Safety}}, pages 225--250. Springer International Publishing, Cham.

\bibitem[{Lakshminarayanan et~al.(2017)Lakshminarayanan, Pritzel, and
  Blundell}]{lakshminarayanan_simple_2017}
Balaji Lakshminarayanan, Alexander Pritzel, and Charles Blundell. 2017.
\newblock \href
  {https://papers.nips.cc/paper_files/paper/2017/hash/9ef2ed4b7fd2c810847ffa5fa85bce38-Abstract.html}
  {Simple and {Scalable} {Predictive} {Uncertainty} {Estimation} using {Deep}
  {Ensembles}}.
\newblock In \emph{Advances in {Neural} {Information} {Processing} {Systems}},
  volume~30. Curran Associates, Inc.

\bibitem[{Li et~al.(2020)Li, Li, Suhara, Doan, and Tan}]{li_deep_2020}
Yuliang Li, Jinfeng Li, Yoshihiko Suhara, AnHai Doan, and Wang-Chiew Tan. 2020.
\newblock \href {https://doi.org/10.14778/3421424.3421431} {Deep {Entity}
  {Matching} with {Pre}-{Trained} {Language} {Models}}.
\newblock \emph{Proceedings of the VLDB Endowment}, 14(1):50--60.

\bibitem[{Liu et~al.(2019)Liu, Ott, Goyal, Du, Joshi, Chen, Levy, Lewis,
  Zettlemoyer, and Stoyanov}]{liu_roberta_2019}
Yinhan Liu, Myle Ott, Naman Goyal, Jingfei Du, Mandar Joshi, Danqi Chen, Omer
  Levy, Mike Lewis, Luke Zettlemoyer, and Veselin Stoyanov. 2019.
\newblock \href {https://doi.org/10.48550/arXiv.1907.11692} {{RoBERTa}: {A}
  {Robustly} {Optimized} {BERT} {Pretraining} {Approach}}.
\newblock ArXiv:1907.11692 [cs].

\bibitem[{Mukhoti et~al.(2020)Mukhoti, Kulharia, Sanyal, Golodetz, Torr, and
  Dokania}]{mukhoti_calibrating_2020}
Jishnu Mukhoti, Viveka Kulharia, Amartya Sanyal, Stuart Golodetz, Philip Torr,
  and Puneet Dokania. 2020.
\newblock \href
  {https://proceedings.neurips.cc/paper/2020/hash/aeb7b30ef1d024a76f21a1d40e30c302-Abstract.html}
  {Calibrating {Deep} {Neural} {Networks} using {Focal} {Loss}}.
\newblock In \emph{Advances in {Neural} {Information} {Processing} {Systems}},
  volume~33, pages 15288--15299. Curran Associates, Inc.

\bibitem[{Méray et~al.(2007)Méray, Reitsma, Ravelli, and
  Bonsel}]{meray_probabilistic_2007}
Nora Méray, Johannes~B. Reitsma, Anita C.~J. Ravelli, and Gouke~J. Bonsel.
  2007.
\newblock \href {https://doi.org/10.1016/j.jclinepi.2006.11.021} {Probabilistic
  {Record} {Linkage} is a {Valid} and {Transparent} {Tool} to {Combine}
  {Databases} {Without} a {Patient} {Identification} {Number}}.
\newblock \emph{Journal of Clinical Epidemiology}, 60(9):883--891.

\bibitem[{Naeini et~al.(2015)Naeini, Cooper, and
  Hauskrecht}]{naeini_obtaining_2015}
Mahdi~Pakdaman Naeini, Gregory Cooper, and Milos Hauskrecht. 2015.
\newblock \href {https://doi.org/10.1609/aaai.v29i1.9602} {Obtaining {Well}
  {Calibrated} {Probabilities} {Using} {Bayesian} {Binning}}.
\newblock \emph{Proceedings of the AAAI Conference on Artificial Intelligence},
  29(1).
\newblock Number: 1.

\bibitem[{Narayan et~al.(2022)Narayan, Chami, Orr, and Ré}]{narayan_can_2022}
Avanika Narayan, Ines Chami, Laurel Orr, and Christopher Ré. 2022.
\newblock \href {https://doi.org/10.14778/3574245.3574258} {Can {Foundation}
  {Models} {Wrangle} {Your} {Data}?}
\newblock \emph{Proceedings of the VLDB Endowment}, 16(4):738--746.

\bibitem[{Niculescu-Mizil and Caruana(2005)}]{niculescu-mizil_predicting_2005}
Alexandru Niculescu-Mizil and Rich Caruana. 2005.
\newblock \href {https://doi.org/10.1145/1102351.1102430} {Predicting {Good}
  {Probabilities} with {Supervised} {Learning}}.
\newblock In \emph{Proceedings of the 22nd {International} {Conference} on
  {Machine} learning}, {ICML} '05, pages 625--632, New York, NY, USA.
  Association for Computing Machinery.

\bibitem[{Peeters et~al.(2020)Peeters, Bizer, and
  Glavas}]{peeters_intermediate_2020}
R.~Peeters, Christian Bizer, and Goran Glavas. 2020.
\newblock \href
  {https://www.semanticscholar.org/paper/Intermediate-Training-of-BERT-for-Product-Matching-Peeters-Bizer/6e177525526925dd30a860bbfc1bbad0c5d42421}
  {Intermediate {Training} of {BERT} for {Product} {Matching}}.
\newblock In \emph{{CEUR} {Workshop} {Proceedings}}, volume 2726, pages 1--2,
  Aachen. Piai, Federico.

\bibitem[{Peeters and Bizer(2021)}]{peeters_dual-objective_2021}
Ralph Peeters and Christian Bizer. 2021.
\newblock \href {https://doi.org/10.14778/3467861.3467878} {Dual-{Objective}
  {Fine}-{Tuning} of {BERT} for {Entity} {Matching}}.
\newblock \emph{Proceedings of the VLDB Endowment}, 14(10):1913--1921.

\bibitem[{Peeters and Bizer(2022)}]{peeters_supervised_2022}
Ralph Peeters and Christian Bizer. 2022.
\newblock \href {https://doi.org/10.1145/3487553.3524254} {Supervised
  {Contrastive} {Learning} for {Product} {Matching}}.
\newblock In \emph{Companion {Proceedings} of the {Web} {Conference} 2022},
  {WWW} '22, pages 248--251, New York, NY, USA. Association for Computing
  Machinery.

\bibitem[{Peeters and Bizer(2023)}]{peeters_using_2023}
Ralph Peeters and Christian Bizer. 2023.
\newblock \href {https://doi.org/10.1007/978-3-031-42941-5_20} {Using {ChatGPT}
  for {Entity} {Matching}}.
\newblock In \emph{New {Trends} in {Database} and {Information} {Systems}},
  pages 221--230, Cham. Springer Nature Switzerland.

\bibitem[{Peeters and Bizer(2024)}]{peeters_entity_2024}
Ralph Peeters and Christian Bizer. 2024.
\newblock \href {https://doi.org/10.48550/arXiv.2310.11244} {Entity {Matching}
  using {Large} {Language} {Models}}.
\newblock ArXiv:2310.11244 [cs].

\bibitem[{Platt(1999)}]{platt_probabilistic_1999}
John Platt. 1999.
\newblock Probabilistic {Outputs} for {Support} {Vector} {Machines} and
  {Comparisons} to {Regularized} {Likelihood} {Methods}.
\newblock \emph{Advances in Large Margin Classifiers}, 10(3):61--74.

\bibitem[{Rahaman and Thiery(2021)}]{rahaman_uncertainty_2021}
Rahul Rahaman and Alexandre~H. Thiery. 2021.
\newblock Uncertainty {Quantification} and {Deep} {Ensembles}.
\newblock \emph{Advances in {Neural} {Information} {Processing} {Systems}},
  34:20063--20075.

\bibitem[{Sanh et~al.(2020)Sanh, Debut, Chaumond, and
  Wolf}]{sanh_distilbert_2020}
Victor Sanh, Lysandre Debut, Julien Chaumond, and Thomas Wolf. 2020.
\newblock \href {https://doi.org/10.48550/arXiv.1910.01108} {{DistilBERT}, a
  {Distilled} {Version} of {BERT}: {Smaller}, {Faster}, {Cheaper} and
  {Lighter}}.
\newblock ArXiv:1910.01108 [cs].

\bibitem[{Sankararaman et~al.(2022)Sankararaman, Wang, and
  Fang}]{sankararaman_bayesformer_2022}
Karthik~Abinav Sankararaman, Sinong Wang, and Han Fang. 2022.
\newblock \href {https://doi.org/10.48550/arXiv.2206.00826} {{BayesFormer}:
  {Transformer} with {Uncertainty} {Estimation}}.
\newblock ArXiv:2206.00826 [cs].

\bibitem[{Vaswani et~al.(2017)Vaswani, Shazeer, Parmar, Uszkoreit, Jones,
  Gomez, Kaiser, and Polosukhin}]{vaswani_attention_2017}
Ashish Vaswani, Noam Shazeer, Niki Parmar, Jakob Uszkoreit, Llion Jones,
  Aidan~N Gomez, Ł~ukasz Kaiser, and Illia Polosukhin. 2017.
\newblock \href
  {https://papers.nips.cc/paper_files/paper/2017/hash/3f5ee243547dee91fbd053c1c4a845aa-Abstract.html}
  {Attention is {All} you {Need}}.
\newblock In \emph{Advances in {Neural} {Information} {Processing} {Systems}},
  volume~30. Curran Associates, Inc.

\bibitem[{Xiao and Wang(2019)}]{xiao_quantifying_2019}
Yijun Xiao and William~Yang Wang. 2019.
\newblock \href {https://doi.org/10.1609/aaai.v33i01.33017322} {Quantifying
  {Uncertainties} in {Natural} {Language} {Processing} {Tasks}}.
\newblock In \emph{Proceedings of the {AAAI} {Conference} on {Artificial}
  {Intelligence}}, volume~33 of \emph{{AAAI}'19/{IAAI}'19/{EAAI}'19}, pages
  7322--7329, Honolulu, Hawaii, USA. AAAI Press.

\bibitem[{Xiao et~al.(2022)Xiao, Liang, Bhatt, Neiswanger, Salakhutdinov, and
  Morency}]{xiao_uncertainty_2022}
Yuxin Xiao, Paul~Pu Liang, Umang Bhatt, Willie Neiswanger, Ruslan
  Salakhutdinov, and Louis-Philippe Morency. 2022.
\newblock \href {https://doi.org/10.18653/v1/2022.findings-emnlp.538}
  {Uncertainty {Quantification} with {Pre}-trained {Language} {Models}: {A}
  {Large}-{Scale} {Empirical} {Analysis}}.
\newblock In \emph{Findings of the {Association} for {Computational}
  {Linguistics}: {EMNLP} 2022}, pages 7273--7284, Abu Dhabi, United Arab
  Emirates. Association for Computational Linguistics.

\bibitem[{Yang et~al.(2019)Yang, Dai, Yang, Carbonell, Salakhutdinov, and
  Le}]{yang_xlnet_2019}
Zhilin Yang, Zihang Dai, Yiming Yang, Jaime Carbonell, Russ~R Salakhutdinov,
  and Quoc~V Le. 2019.
\newblock \href
  {https://proceedings.neurips.cc/paper_files/paper/2019/file/dc6a7e655d7e5840e66733e9ee67cc69-Paper.pdf}
  {{XLNet}: {Generalized} {Autoregressive} {Pretraining} for {Language}
  {Understanding}}.
\newblock In \emph{Advances in {Neural} {Information} {Processing} {Systems}},
  volume~32. Curran Associates, Inc.

\end{thebibliography}
